%% file: main.tex
\documentclass{article}

\PassOptionsToPackage{numbers, compress}{natbib}

\usepackage[final]{neurips_2022}

\usepackage[utf8]{inputenc} 
\usepackage[T1]{fontenc}    
\usepackage{hyperref}       
\usepackage{url}            
\usepackage{booktabs}       
\usepackage{amsfonts}       
\usepackage{nicefrac}       
\usepackage{microtype}      
\usepackage{xcolor}         

\usepackage{xspace}
\usepackage{multirow}
\usepackage{graphicx}
\usepackage{adjustbox}
\usepackage{makecell}
\usepackage{mathtools}
\usepackage{wrapfig}

\newcommand{\ie}{{\it i.e.}}
\newcommand{\eg}{{\it e.g.}}
\newcommand{\etal}{{\it et al.}\xspace}

\usepackage{enumitem}
\usepackage{amsmath}
\usepackage{amssymb}
\usepackage{amsthm, thmtools}

\newtheorem{assumption}{Assumption}

\newtheorem{lem}{Lemma}

\usepackage{epsfig}
\usepackage{multirow}
\usepackage{makecell}
\usepackage{tabularx}
\usepackage{algorithm} 
\usepackage{algpseudocode} 
\usepackage{pifont}

\DeclareMathOperator\supp{supp}
\newcommand{\indexset}{\mathcal{I}}

\newcommand{\std}[1]{\tiny{$\pm$#1}}
\newcommand{\cell}[2][1]{#1\std{#2}}
\newcommand{\bcell}[2][1]{\textbf{#1\std{#2}}}
\newcommand{\lcell}[2][1]{\underline{#1\std{#2}}}

\title{Mining Multi-Label Samples from Single Positive Labels}

\newcommand*{\affaddr}[1]{#1}
\newcommand*{\affmark}[1][*]{\textsuperscript{#1}}
\newcommand*{\email}[1]{\texttt{#1}}
\author{
Youngin Cho*\affmark[1]\quad Daejin Kim*\affmark[1,2]\quad Mohammad Azam Khan\affmark[1]\quad Jaegul Choo\affmark[1]\\
\affaddr{\affmark[1]KAIST AI}\quad \affaddr{\affmark[2]NAVER WEBTOON AI}\\
\small\email{\{choyi0521,kiddj,azamkhan,jchoo\}@kaist.ac.kr}
}

\begin{document}

\maketitle
\def\thefootnote{*}\footnotetext{Equal contribution}\def\thefootnote{\arabic{footnote}}

\input{00_Abstract}
\input{01_Introduction}
\input{02_Related_Works}
\input{03_Method}
\input{04_Experiments}
\input{05_Limitations}
\input{06_Conclusion}


\begin{ack}
This work was supported by the Institute of Information \& communications Technology Planning \& Evaluation (IITP) grant funded by the Korean government (MSIT) (No. 2019-0-00075, Artificial Intelligence Graduate School Program (KAIST) and No. 2021-0-01778, Development of human image synthesis and discrimination technology below the perceptual threshold). This work was also supported by the National Research Foundation of Korea (NRF) grant funded by the Korean government (MSIT) (No. NRF-2022R1A2B5B02001913).
\end{ack}

\bibliographystyle{unsrt}
\bibliography{references}

\newpage
\section*{Checklist}

\begin{enumerate}

\item For all authors...
\begin{enumerate}
  \item Do the main claims made in the abstract and introduction accurately reflect the paper's contributions and scope?
    \answerYes{}
  \item Did you describe the limitations of your work?
    \answerYes{See Section~\ref{sec:limitations}.}
  \item Did you discuss any potential negative societal impacts of your work?
    \answerYes{See Appendix.}
  \item Have you read the ethics review guidelines and ensured that your paper conforms to them?
    \answerYes{}
\end{enumerate}

\item If you are including theoretical results...
\begin{enumerate}
  \item Did you state the full set of assumptions of all theoretical results?
    \answerYes{}
        \item Did you include complete proofs of all theoretical results?
    \answerYes{See Appendix.}
\end{enumerate}

\item If you ran experiments...
\begin{enumerate}
  \item Did you include the code, data, and instructions needed to reproduce the main experimental results (either in the supplemental material or as a URL)?
    \answerYes{We provide the codes in the supplemental material.}
  \item Did you specify all the training details (e.g., data splits, hyperparameters, how they were chosen)?
    \answerYes{See Appendix.}
        \item Did you report error bars (e.g., with respect to the random seed after running experiments multiple times)?
    \answerYes{We reported the standard deviation in the experimental results.}
        \item Did you include the total amount of compute and the type of resources used (e.g., type of GPUs, internal cluster, or cloud provider)?
    \answerYes{See Appendix.}
\end{enumerate}

\item If you are using existing assets (e.g., code, data, models) or curating/releasing new assets...
\begin{enumerate}
  \item If your work uses existing assets, did you cite the creators?
    \answerYes{}
  \item Did you mention the license of the assets?
    \answerNo{We used publicly available benchmark datasets: MNIST, FMNIST, CIFAR10, CelebA, and CelebA-HQ.}
  \item Did you include any new assets either in the supplemental material or as a URL?
    \answerYes{We provide the codes in the supplemental material.}
  \item Did you discuss whether and how consent was obtained from people whose data you're using/curating?
    \answerYes{The benchmark datasets used in our paper are widely used for research purpose.}
  \item Did you discuss whether the data you are using/curating contains personally identifiable information or offensive content?
    \answerYes{The benchmark datasets used in our paper are widely used by others and do not contain any offensive contents.}
\end{enumerate}

\item If you used crowdsourcing or conducted research with human subjects...
\begin{enumerate}
  \item Did you include the full text of instructions given to participants and screenshots, if applicable?
    \answerNA{}
  \item Did you describe any potential participant risks, with links to Institutional Review Board (IRB) approvals, if applicable?
    \answerNA{}
  \item Did you include the estimated hourly wage paid to participants and the total amount spent on participant compensation?
    \answerNA{}
\end{enumerate}

\end{enumerate}


\input{Appendix}

\end{document}

%% file: 00_Abstract.tex
\begin{abstract}
Conditional generative adversarial networks (cGANs) have shown superior results in class-conditional generation tasks. To simultaneously control multiple conditions, cGANs require multi-label training datasets, where multiple labels can be assigned to each data instance. Nevertheless, the tremendous annotation cost limits the accessibility of multi-label datasets in real-world scenarios. Therefore, in this study we explore the practical setting called the \emph{single positive setting}, where each data instance is annotated by only one positive label with no explicit negative labels. To generate multi-label data in the single positive setting, we propose a novel sampling approach called single-to-multi-label (S2M) sampling, based on the Markov chain Monte Carlo method. As a widely applicable “add-on” method, our proposed S2M sampling method enables existing unconditional and conditional GANs to draw high-quality multi-label data with a minimal annotation cost. Extensive experiments on real image datasets verify the effectiveness and correctness of our method, even when compared to a model trained with fully annotated datasets.
\end{abstract}

%% file: 01_Introduction.tex
\section{Introduction}
\label{sec:introduction}

Since being proposed by Goodfellow \etal~\cite{GoodfellowPMXWOCB14}, generative adversarial networks (GANs) have gained much attention due to their realistic output in a wide range of applications, \eg, image synthesis~\cite{BrockDS19, KarrasALL18, Park0WZ19}, image translation~\cite{LiuBK17, ZhuPIE17, Isola2017, ChoiCKH0C18}, and data augmentation~\cite{ShrivastavaGG16, abs-1810-10863}.
As an advanced task, generating images from a given condition has been achieved by conditional GANs (cGANs) and their variants~\cite{MirzaO14, OdenaOS17}.
To reflect the nature of real data where each data instance can belong to multiple classes, multi-label datasets have been introduced in the applications of cGANs~\cite{ChoiCKH0C18, LinWCCL19}.
In a multi-label dataset, each data instance can be specified with multiple attributes. 
For example, in a multi-label facial image dataset, each image is labeled for entire classes such as \emph{Black-hair}, \emph{Smile}, and \emph{Male}.
The label for each class is given as a binary value and indicates the presence or absence of the corresponding attribute.
Despite its usefulness, as claimed in earlier studies~\cite{LinMBHPRDZ14, GuptaDG19, ColeALPMJ21}, access to multi-label datasets is severely limited in practice due to the difficulty of annotating all attributes.
Under these circumstances, a weakly annotated dataset is used as an alternative for cGANs to reduce the annotation cost.

\input{Figures/Figure_Concepts}

In this paper, we introduce the \emph{single positive setting}~\citep{ColeALPMJ21}, originally proposed for classification tasks, to conditional generation. 
Each data instance has a label indicating only the presence of one attribute (\ie, a single positive label) in this setting, and the presence of the rest of the attributes remains unknown.
For instance, in a facial image dataset, all attributes are fully specified in a multi-label setting (\eg, \emph{Smiling black-haired man}) whereas only one attribute is specified in the single positive setting (\eg, \emph{Black-hair}).
The single positive setting allows us not only to reduce the annotation cost, but also to model the intrinsic relationships among classes.

To generate multi-label data using only single positive labels, we consider two types of combinatorial classes from two classes, $A$ and $B$, each corresponding to a single positive label: (1) \emph{overlapping class}, where the data instances belong to both classes ($A \cap B$), and (2) \emph{non-overlapping class}, where the data instances belong to only one of the classes ($A \setminus B$ or $B \setminus A$).
Figure~\ref{fig:method_concepts} shows examples of overlapping class and non-overlapping classes when two types of single positive labels are given.
By extending this concept, we can consider combinatorial classes, where data instances belong to certain classes but not to the rest. We denote these classes as \emph{joint classes}.
By accessing a joint class, we can generate multi-label samples using only single positive labels.
Ideally, we can represent all possible label combinations with at least one positive label.

Several attempts have been made to consider such a joint class in generation tasks.
Specially designed generative models such as GenPU~\cite{HouCLZ18} and RumiGAN~\cite{AsokanS20} were proposed to generate samples from one class while excluding another class. However, these studies deal with only two classes without considering the overlapping class. 
Recently, Kaneko \etal~\cite{KanekoUH19} proposed CP-GAN to capture between-class relationships in conditional generation.
To generate images belonging to $n$ classes, they equally assign $1/n$ as the class posterior for each selected class. This indicates that CP-GAN generates images that have an equal probability of being classified as each class.
However, because an image in the real world that belongs to multiple classes does not have equal class posteriors, sample diversity is lost.
Consequently, we focus on a sampling approach to precisely draw samples from complex distributions that are difficult to directly sample.
In recent GANs studies, sampling approaches~\cite{AzadiODGO19, TurnerHFSY19} employed the rejection sampling or Markov chain Monte Carlo method to obtain realistic data.
These approaches adopt the sampling process as the post-processing method of GANs and filter generated images using the pretrained classifiers.

In line with these studies, we propose a novel sampling framework called \emph{single-to-multi-label (S2M) sampling} to correctly generate data from both overlapping and non-overlapping classes.
We newly propose a tractable formulation to estimate the conditional density of joint classes.
Concretely, we employ several classification networks to estimate the target conditional density and apply the Markov chain Monte Carlo method to pretrained unconditional and conditional GANs.
In Figure~\ref{fig:method_comparison}, we depict the conceptual difference against the existing approaches ($1^\textrm{st}$ row) and provide the empirical results on a 1D Gaussian example ($2^\textrm{nd}$ row).
As S2M sampling performs at the inference time of generation, it fully preserves the image quality and eliminates the need for changing the objective functions and architectures.
We validate the effectiveness of our method compared to the existing models on diverse datasets such as MNIST, CIFAR-10 and CelebA. 
To the best of our knowledge, our approach is the first sampling framework that generates multi-label data from single positive labels. Our contributions can be summarized as follows:
\begin{itemize}[leftmargin=0.25in]
    \setlength\itemsep{0.1em}
    \item We introduce the single positive setting in the conditional generation task and provide the theoretical framework for estimating conditional densities of joint classes.
    \item We propose a novel sampling framework based on the Markov chain Monte Carlo method for generating multi-label data from single positive labels.
    \item Through extensive experiments, we show that our proposed S2M sampling method correctly draws multi-label data while fully preserving the quality of generated images.
\end{itemize}

\input{Figures/Figure_Method_Comparison}

%% file: Figures/Figure_Concepts.tex
\begin{figure}[t!]
\begin{center}
    \includegraphics[width=1.0\linewidth]{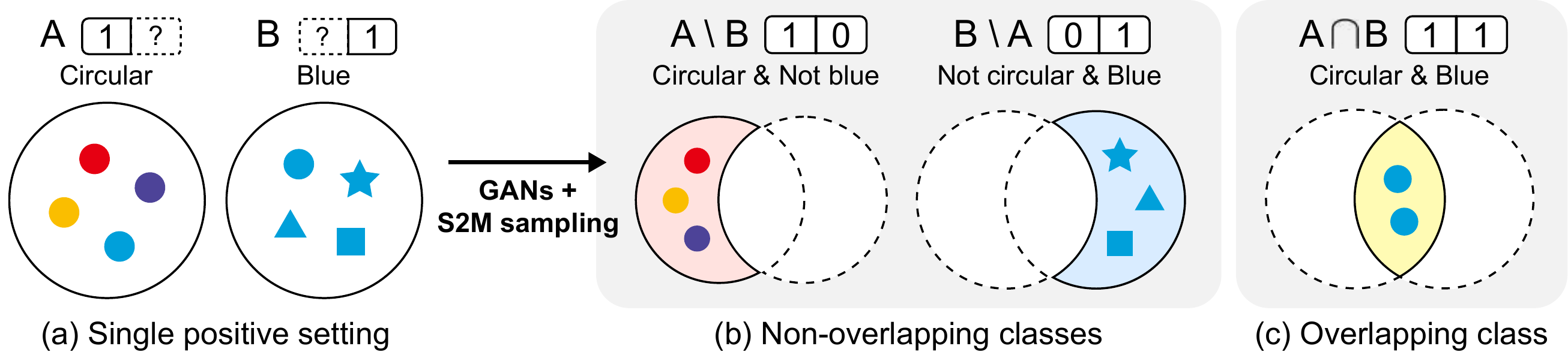}
\end{center}
    \caption{Illustration of joint classes where two attributes (circular, blue) are given. (a) Only one of the attributes is specified in the single positive setting.
    From classes $A$ and $B$, our proposed S2M sampling method can draw samples from three joint classes: (b) two non-overlapping classes ($A\setminus B$ and $B\setminus A$) and (c) one overlapping class ($A \cap B$).}
\label{fig:method_concepts}
\end{figure}

%% file: Figures/Figure_Method_Comparison.tex
\begin{figure*}[t!]
\begin{center}
    \includegraphics[width=1.0\linewidth]{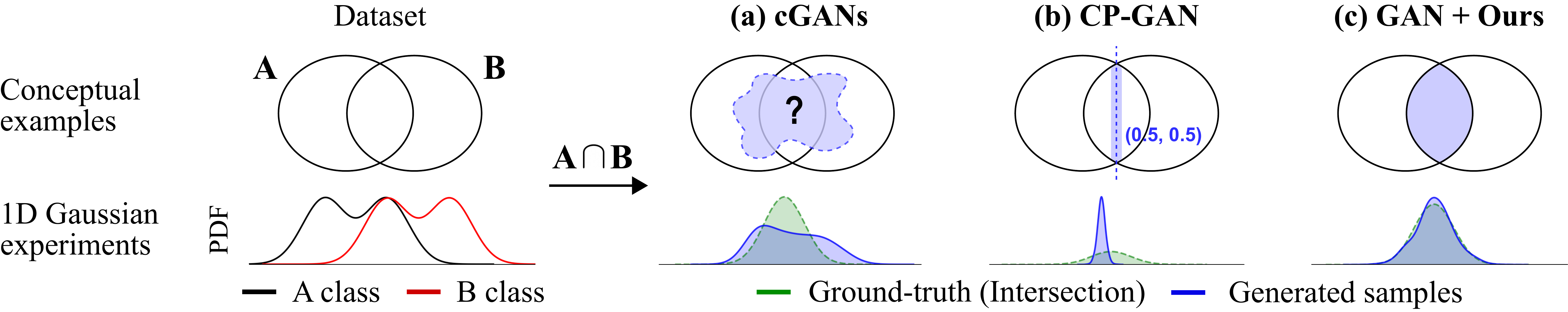}
\end{center}
\vspace{-0.1in}
    \caption{cGANs, CP-GAN and our method are compared in a class-overlapping case. 1D Gaussian examples consists of two classes of one-dimensional Gaussian mixtures with one common mode, and each method attempts to generate samples in the overlapping region. For cGANs and CP-GAN, an equal value of $0.5$ is provided as labels for the two classes. (a) It is not clear how cGANs obtain samples of the class. (b) CP-GAN draws samples from the narrow region. (c) GAN with S2M sampling draws samples accurately without sacrificing diversity.}
\vspace{-0.1in}
\label{fig:method_comparison}
\end{figure*}

%% file: 02_Related_Works.tex
\section{Related Work}

\textbf{Conditional GANs.}
The aim of conditional GANs (cGANs)~\citep{MirzaO14} is to model complex distributions and control data generation by reflecting the label input. Various studies of cGANs have made significant advances in class-conditional image generation by introducing an auxiliary classifier~\citep{OdenaOS17, GongXLZB19}, modifying the architecture~\citep{MiyatoK18, BrockDS19}, and applying metric learning~\citep{KangP20}. 
In a weakly-supervised setting, GenPU~\citep{HouCLZ18} and RumiLSGAN~\citep{AsokanS20} specify only two classes and draw samples that belong to one class but not the other.
CP-GAN~\citep{KanekoUH19} learns to draw samples conditioned on the probability output of the classifier. Given that this model tends to draw samples on a limited region of the data space, it is challenging to ensure a variety of samples as shown in Figure~\ref{fig:method_comparison}. In contrast, we propose a sampling method that draws multi-label samples without sacrificing diversity.

\textbf{Sampling in GANs.} Sampling methods are used to improve the sample quality in GANs. Discriminator rejection sampling~\citep{AzadiODGO19} uses the scheme of rejection sampling and takes samples close to real data by estimating the density ratio with the discriminator. In addition, Metropolis-Hastings GAN~\citep{TurnerHFSY19} adopts the Markov chain Monte Carlo (MCMC) method and calibrates the discriminator to improve the sample quality in a high-dimensional data space. 
Discriminator driven latent sampling~\citep{CheZSLPCB20} uses the MCMC method in the latent space of GANs to draw realistic samples efficiently. 
GOLD estimator~\citep{MoKKCS19} uses a sampling algorithm to improve the quality of images for class-conditional generation. While previous studies focus on improving the sample quality, our S2M sampling method aims to draw multi-label samples while also improving sample quality.

%% file: 03_Method.tex
\section{Methods}
\label{sec:method}
\subsection{Problem Setting}

Let $x\in X$ be a data point as a random variable and let $y_{1:n} \in \{0, 1\}^n$ denote its corresponding multi-labels as binary random variables. Here, for every $k$, $y_k=1$ indicates that $x$ is contained in the $k$-th class while $y_k=0$ indicates that $x$ is not.
We consider two index sets, an intersection index set $I$ and a difference index set $J$, so that the pair $(I,J)$ can be used as an index to indicate the class, where data points contained in all classes indicated by $I$ but excluded from all classes indicated by $J$. 
Let $\indexset$ be a collection of all possible pairs of $I$ and $J$, defined as
\begin{equation}
\indexset=\{(I,J)\in \mathcal{P}(N) \times \mathcal{P}(N):I\neq \emptyset, I \cap J = \emptyset\},
\end{equation}
where $N=\{1, 2, ..., n\}$ is a finite index set of all classes and $\mathcal{P}(N)$ is the power set of $N$. That is, the intersection index set indicates at least one class and is distinct from the difference index set. Especially for $I\cup J = N$, the class indicated by $(I, J)$ is called the $\emph{joint class}$.

Let $p(x,y_1, y_2, ..., y_n)$ be the joint probability density function, and let $p_{data}(x, c)$ be the joint density of an observed data point $x\in X$ and a class label $c\in N$ such that $p_{data}(x|c)=p(x|y_c=1)$.
Given the class priors $p_{data}(c), \pi_c := p(y_c=1)$ and samples drawn from the class-conditional density $p_{data}(x|c)$ for each $c=1,2,...,n$, our goal is to draw samples from the conditional density
\begin{equation}
\label{eq:target_density}
    p_{(I,J)}(x) := p(x|\forall i \in I, \forall j \in J,  y_{i}=1, y_{j}=0),
\end{equation}
for $(I, J)\in \indexset$ and $\pi_{(I,J)}:=p(\forall i \in I, \forall j \in J,  y_{i}=1, y_{j}=0)>0$. 
In this work, we propose adding a mild constraint which will allow our sampling algorithm to derive the target density.
\begin{assumption}
\label{as:distinct_class_separability}
For every $i, j\in N$ such that $i\neq j$, if $p(y_i=1, y_j=0)>0$ and $p(y_j=1, y_i=0)>0$, then $\supp p(x|y_i=1, y_j=0) \cap \supp p(x|y_j=1, y_i=0) = \emptyset.$
\end{assumption}
Assumption~\ref{as:distinct_class_separability} states that no data points are likely to be assigned two mutually exclusive labels, which can be naturally assumed in many practical situations. Figure~\ref{fig:high_level_explanation} illustrates our problem setting.

\input{Figures/Figure_Method_Introduction}

\subsection{Mining Multi-Label Data with S2M Sampling}
\label{sec:s2m_sampling}

Naturally, a question may arise as to whether the supervision given to us is sufficient to obtain multi-label samples.
To gain insight into this, we derive a useful theorem which provides an alternative formulation for the target density~(\ref{eq:target_density}).

\begin{restatable}{thm}{universe}
\label{thm:equivalence}
Let $\{f_{(I,J)}:X\to[0,\infty)\}_{(I,J)\in\indexset}$ be an indexed family of non-negative measurable functions on $X$, and let $f_k\coloneqq f_{(\{k\},\emptyset)}$.
Then, the following conditions hold:
\begin{enumerate}[label=(\alph*),noitemsep,topsep=0pt]
\item $\forall (I, J)\in \indexset, f_{(I, J)} = \sum_{S:I\subseteq S, J\subseteq N\setminus S} f_{(S,N\setminus S)}$
\item $\forall i, j\in N \text{ s.t. } i\neq j, \supp f_{(\{i\},\{j\})} \cap \supp f_{(\{j\},\{i\})} = \emptyset$
\end{enumerate}
if and only if, for every $(I, J)\in \indexset$,
\vspace{-0.05in}
\begin{equation}
    f_{(I, J)}=
    \begin{cases}
    \left(\min_{i\in I}f_i - \max_{j\in J}f_j\right)^+ &\mbox{if } J\neq \emptyset \\
    \min_{i\in I}f_i & \mbox{otherwise}
    \end{cases},
\end{equation}
\vspace{-0.05in}
where $(\cdot)^+$ represents the positive part.
\end{restatable}
\vspace{-0.1in}
\begin{proof}
Please refer to Appendix~\ref{appendix:proof}.
\end{proof}
\vspace{-0.1in}
Let $f_{(I,J)}(x)=p(x, \forall i \in I, \forall j \in J,  y_{i}=1, y_{j}=0)$ for every $(I,J) \in \indexset$. Then, both $(a)$ and $(b)$ in Theorem~\ref{thm:equivalence} hold by the Assumption~\ref{as:distinct_class_separability}. According to the Theorem~\ref{thm:equivalence}, if $\pi_{(I,J)}>0$,
\begin{equation}
\label{eq:altform}
\begin{aligned}
    p_{(I,J)}(x) &= \pi_{(I,J)}^{-1} (\min\{\pi_i p(x|y_i=1): i\in I\} - \max\{\pi_j p(x|y_j=1):j\in J\}\cup\{0\})^+.
\end{aligned}
\end{equation}
The alternative formula~($\ref{eq:altform}$) shows that the density of the joint class can be derived from the class-conditional densities of the single positive labels.
Despite their clear relationship, training a generator that can model the density $p_{(I,J)}$ is a non-trivial problem since the formula consists of several implicitly defined conditional densities, class priors, and variable sets. To address this issue, we propose the application of sampling approaches upon existing GANs. Interestingly, our sampling framework called S2M sampling allows not only for samples to be drawn from the target density, but also the modification of $I$, $J$, and class priors at inference time.
The rest of this section introduces the main approaches of our S2M sampling method.

\textbf{Density Ratio Estimation.} 
We utilize several classification networks to compute implicitly defined density ratios.
For simplicity, we denote $G$ as a pretrained generator for both unconditional and conditional GANs. $G$ produces data $x$ by taking a latent $z$ and a class label $c$ for class-conditional generation and only $z$ for unconditional generation.
We consider three classifiers $D_v, D_r$, and $D_f$ which are obtained by minimizing $\mathcal{L}_v, \mathcal{L}_r$, and $\mathcal{L}_f$, respectively, \ie, 
\begin{equation}
\begin{aligned}
    \mathcal{L}_v = &-\mathbb{E}_{(x,c)\sim p_{data}(x,c)}[\log D_v(x)] - \mathbb{E}_{x\sim p_G(x)}[\log (1- D_v(x))], \\
    \mathcal{L}_r = &-\mathbb{E}_{(x,c)\sim p_{data}(x,c)}[\log D_r(c|x)], \
    \mathcal{L}_f = -\mathbb{E}_{(x,c)\sim p_G(x,c)}[\log D_f(c|x)],
\end{aligned}
\end{equation}
where $p_G$ is the distribution of the generated samples by $G$. The optimal classifiers trained by these losses $D_v^\ast$, $D_r^\ast$, and $D_f^\ast$ satisfy the following equations: $D_v^\ast(x) = p_{data}(x)/(p_{data}(x) + p_G(x)), D_r^\ast(c|x) = p(x|y_c=1) p_{data}(c)/p_{data}(x), D_f^\ast(c|x) = p_G(x|c) p_G(c)/p_G(x).$
From $D_v^\ast$, $D_r^\ast$, and $D_f^\ast$, we can access the density ratios $p_{data}(x)/p_G(x)$, $p(x|y_c=1)/p_{data}(x)$, and $p_G(x|c)/p_G(x)$ which will be used to compute the acceptance probability of the MCMC method.

\textbf{S2M Sampling for Unconditional GANs.} We apply Metropolis-Hastings (MH) independence sampling~\citep{luke1994markov, TurnerHFSY19} to draw samples from the complex target distribution $p_t$. 
The MH algorithm uses a Markov process where each transition from a current state $x_k$ to the next state $x_{k+1}$ is made by an accept-reject step.
At each step of MH independent sampling, a new proposal $x'$ is sampled from a proposal distribution $q(x'|x)=q(x')$ and is then accepted with probability $\alpha(x',x)$ which is given by $\alpha(x',x)=\min\left(1, \frac{p_t(x')q(x)}{{p_t(x)q(x')}}\right)$.
We set $x_{k+1} = x'$ if $x'$ is accepted, and $x_{k+1} = x_k$ otherwise. Under mild conditions, the chain $x_{1:K}$ converges to the unique stationary distribution $p_t$ as $K\rightarrow \infty$\footnote{For example, the chain of samples is uniformly ergodic if $p_t/q$ is bounded~\cite{1033066201}.}

Let $G$ be a generator where the support of $p_G$ contains that of $p_{(I, J)}$. To draw samples from $p_{(I, J)}$, we let $p_t = p_{(I, J)}$ and take multiple samples from $G$ as independent proposals, i.e. $q(x'|x)=p_{G}(x')$. Then, the desired acceptance probability $\alpha(x', x)$ can be calculated using $D_v^\ast$ and $D_r^\ast$:
\begin{equation}
\label{eq:alpha}
\begin{aligned}
r_{(I, J)}(x) &\coloneqq \left(\min\left\{\frac{\pi_i}{p_{data}(i)}D_r^\ast(i|x): i\in I\right\} \right.\left.- \max\left\{\frac{\pi_j}{p_{data}(j)}D_r^\ast(j|x):j\in J\right\}\cup\left\{0\right\} \right)^+,\\
    \alpha(x', x) & = \min\left(1, \frac{p_{(I,J)}(x')/p_G(x')}{p_{(I,J)}(x)/p_G(x)}\right)
    = \min\left(1,\frac{r_{(I,J)}(x')({D_v^\ast(x)}^{-1}-1)}{r_{(I,J)}(x)({D_v^\ast(x')}^{-1}-1)}\right).
\end{aligned}
\end{equation}
Different from Turner \etal~\cite{TurnerHFSY19}, the term $r_{(I,J)}(x')/r_{(I,J)}(x)$ is added to the acceptance probability formula to draw multi-label samples. To obtain uncorrelated samples, a sample is taken after a fixed number of iterations for each chain. The sampling approach allows one to control the parameters $I$, $J$, and $\gamma_k \coloneqq \pi_k/p_{data}(k)$ without any additional training of the model. A summary of our S2M sampling algorithm is provided in Appendix~\ref{appendix:s2m_pseudocode}.

\textbf{S2M Sampling for Conditional GANs.}
cGANs can provide a proposal distribution close to the target distribution $p_{(I, J)}$, which greatly increases the sample efficiency of the MCMC method. Let $c$ be a class label such that the support of class-conditional density $p_G(\cdot|c)$ contains that of $p_{(I, J)}$. At each step of the MH algorithm, the proposal $x' \sim q(x'|x)=p_{G}(x'|c)$ is accepted with a probability $\alpha_c(x', x)$. The desired $\alpha_c(x', x)$ can be calculated as 
\begin{equation}
\label{eq:alpha_c}
    \begin{aligned}
    &\alpha_c(x', x)  = \min\left(1, \frac{p_{(I,J)}(x')/p_G(x'|c)}{p_{(I,J)}(x)/p_G(x|c)}\right) = \min\left(1,\frac{r_{(I,J)}(x')D_f^\ast(c|x)({D_v^\ast(x)}^{-1}-1)}{r_{(I,J)}(x)D_f^\ast(c|x')({D_v^\ast(x')}^{-1}-1)}\right).
    \end{aligned}
\end{equation}
That is, the sampling method can be adopted to cGANs by additionally computing $D_f^\ast(c|x)/D_f^\ast(c|x')$.

\textbf{Latent Adaptation in S2M Sampling.}
While our S2M sampling method allows us to draw multi-label samples, the algorithm rejects certain samples through the sampling procedure if the target distribution $p_{(I,J)}$ is significantly different from the generator distribution.
Specifically, an independent MH algorithm takes time inversely proportional to the probability of a generated sample belonging to the target class. 
To alleviate this issue, we propose a technique to improve the sample efficiency from past sampling attempts.
Initially, a certain number of pilot target class samples $x_{1:m}$ are drawn using S2M sampling, and then the corresponding target latent samples $t_{1:m}$; $x_k = G(t_k)$ are obtained. Since the latent samples are nearly restricted to the latent prior of GANs (\eg, Gaussian distribution), the distribution $\tilde{p}_z$ of the target latent can be roughly estimated by fitting a simple probabilistic model using $t_{1:m}$. Let $\tilde{p}_G$ be the newly obtained generator distribution induced by $G$, \ie, $G(z)=x \sim\tilde{p}_G(x)$ for $z \sim \tilde{p}_z(z)$. The sample efficiency can be further improved by using $\tilde{p}_G$ as the proposal distribution in the MH algorithm since $\tilde{p}_G$ is much close to the target distribution.
To run the MH algorithm without retraining the classifiers, we approximate $p_G(G(z))/\tilde{p}_G(G(z)) \approx C \cdot p_z(z)/\tilde{p}_z(z)$\footnote{The equation is derived in the previous latent sampling approaches by the reverse lemma of rejection sampling~\citep{CheZSLPCB20} or the change-of-variables formula~\citep{WangWYL21}.} for some constant $C$. Then, at each step of the MH algorithm, the proposal $x' \sim q(x'|x)=\tilde{p}_G(x')$ is accepted with a probability $\tilde{\alpha}(x', x)$ which can be calculated as
\begin{equation}
\label{eq:alpha_la}
    \begin{aligned}
    &\tilde{\alpha}(x', x)  = \min\left(1, \frac{p_{(I,J)}(x')/\tilde{p}_G(x')}{p_{(I,J)}(x)/\tilde{p}_G(x)}\right) \approx \min\left(1,\frac{r_{(I,J)}(x')({D_v^\ast(x)}^{-1}-1)p_z(z')/\tilde{p}_z(z')}{r_{(I,J)}(x)({D_v^\ast(x')}^{-1}-1)p_z(z)/\tilde{p}_z(z)}\right),
    \end{aligned}
\end{equation}
where $x'=G(z')$ and $x=G(z)$. Here, we can explicitly compute the density ratio between $p_z$ and $\tilde{p}_z$ by letting $\tilde{p}_z$ be a known probability density function such as a Gaussian mixture. 
In practice, if the target class samples rarely appear in the generated samples, one can consider applying latent adaptation repeatedly.
For instance, to draw \emph{Black-haired man} samples, we can first search the latent space of \emph{Man} by using latent adaptation, and then perform it in that space again to search the space of \emph{Black-haired man}.
In terms of time complexity, this is performed more efficiently than searching the space of \emph{Black-haired man} at once. A detailed description of latent adaptation and a discussion about the time complexity are provided in Appendix~\ref{appendix:s2m_latent_adaptation} and Appendix~\ref{appendix:time_complexity}, respectively.

\textbf{Practical Considerations.}
We employ three classifiers $D_v$, $D_r$, and $D_f$, to compute the acceptance probability used in the MCMC method. For better training efficiency of the classifiers, we use shared layers, except for the last linear layer.
Since the classifiers are not optimal in practice, we scale the temperature of classifiers~\citep{GuoPSW17} and adjust $\gamma_k$ to calibrate the sampling algorithm. 
Detailed settings and the ablation study are provided in Appendix~\ref{appendix:exp_details} and Appendix~\ref{appendix:ablation_study}, respectively.

%% file: Figures/Figure_Method_Introduction.tex
\begin{figure}[t!]
\begin{center}
    \includegraphics[width=1.0\linewidth]{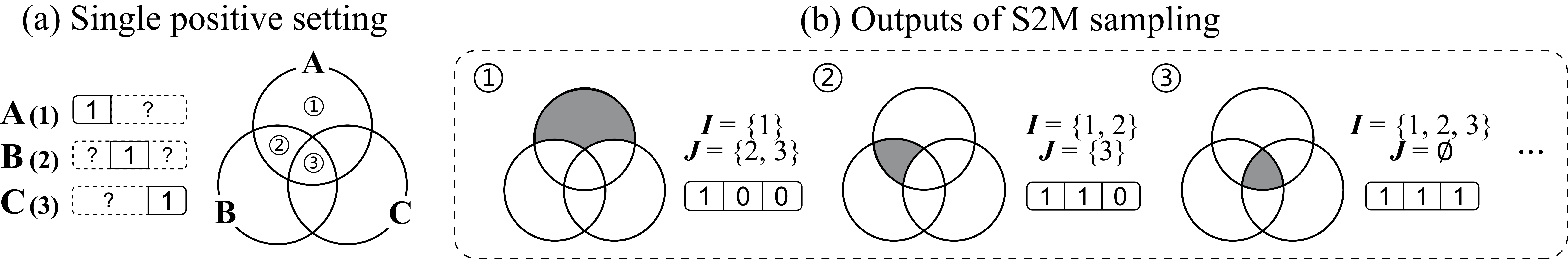}
\end{center}
\vspace{-0.1in}

    \caption{(a) A dataset with single positive labels is given. 
    (b) S2M sampling can draw samples as multi-label data with two index sets: intersection ($I$) and difference ($J$).
    }
\vspace{-0.1in}
\label{fig:high_level_explanation}
\end{figure}

%% file: 04_Experiments.tex
\section{Experiments}

\input{Figures/Figure_Experiments_Settings}

In this section, we validate that our S2M sampling method properly draws samples within the joint classes. Specifically, we mainly consider two cases in the single positive setting: (i) one class is entirely contained in another class, (ii) multiple classes are partially overlapping with each other. 
In the former setting, we verify that S2M sampling can completely filter out samples of a smaller class, and compare our method to the existing models including GenPU~\citep{HouCLZ18} and RumiLSGAN~\citep{AsokanS20} on MNIST and Fashion-MNIST (FMNIST).
In the latter setting, all possible joint classes are assessed with our S2M sampling method.
Subsequently, we evaluate how accurately S2M sampling draws samples in these classes compared to CP-GAN~\citep{KanekoUH19} and the models trained with a fully annotated multi-label dataset.

To verify whether a method can accurately generate samples for each joint class, we evaluate \emph{accuracy} which indicates how many generated images are assigned to the target joint classes by a classifier trained with fully annotated data (\ie, multi-label data). 
Since accuracy itself cannot fully evaluate the distance between the distribution of generated samples and that of real samples, we additionally introduce various metrics to evaluate fidelity (\ie, how realistic the generated samples are) and diversity of generated images: (i) \emph{Inception Score} (IS)~\cite{SalimansGZCRCC16}, (ii) \emph{Fréchet Inception Distance} (FID)~\cite{HeuselRUNH17}, (iii) \emph{improved precision and recall}~\cite{KynkaanniemiKLL19}, and (iv) \emph{density and coverage}~\cite{NaeemOUCY20}. 

FID and IS are metrics to evaluate the fidelity and the diversity of generated samples using the features of the pretrained Inception-V3 network~\cite{SzegedyVISW16}. A lower FID and a higher IS indicate higher fidelity and diversity of generated samples. 
Precision and density measure the ratio of generated samples that falls within the manifold of real samples, so that they evaluate the fidelity of generated samples. Contraily, recall and coverage measure the ratio of real samples that falls within the manifold of the generated samples, and thus evaluating their diversity. That is, a higher value of these metrics indicates that the generated samples have a distribution similar to the real one.

Since several metrics for fidelity and diversity are not applicable for non-ImageNet-like images (\eg, MNIST and FMNIST), we instead use the FID score computed from activations of LeNet5~\citep{726791} and denote this as $\textrm{FID}^\dagger$. 
The test set is used as a reference dataset for evaluating FID.
All quantitative results are averaged over three independent trials, and the standard deviation is denoted by subscripts. The qualitative results are randomly sampled in all experiments. Detailed experimental settings, such as architectures and hyperparameters, are provided in Appendix~\ref{appendix:exp_details}. 


\subsection{Sampling for Classes with Inclusion}
\label{sec:exp_mnist}

\textbf{Experimental Settings.}
We consider a special case of our problem setting, where one class is entirely contained in another class ($B \subset A$) as shown in Figure~\ref{fig:exp_settings} (a). This setting is similar to the positive-unlabeled setting~\citep{Denis98, DenisGL05} where the positive data is included in the unlabeled data. Under this constraint, GenPU~\citep{HouCLZ18}, RumiLSGAN~\citep{AsokanS20}, and CP-GAN~\citep{KanekoUH19} can be used to generate samples from the non-overlapping class ($A \setminus B$).
Along with these models, we attempt to draw samples from the non-overlapping class by adopting S2M sampling to unconditional WGAN-GP~\citep{WeiGL0W18}.
These experiments were conducted in three settings (See Figure~\ref{fig:exp_settings}): (i) \textbf{MNIST-3/5} ($A$=\{3, 5\}, $B$=\{3\}), (ii) \textbf{MNIST-Even} ($A$=\{0, 1, 2, 3, 4, 5, 6, 7, 8, 9\}, $B$=\{1, 3, 5, 7, 9\}), and (iii) \textbf{FMNIST-Even} ($A$=$\{0_\textrm{T-shirt/Top}$, $1_\textrm{Trouser}$, $2_\textrm{Pullover}$, $3_\textrm{Dress}$, $4_\textrm{Coat}$, $5_\textrm{Sandal}$, $6_\textrm{Shirt}$, $7_\textrm{Sneaker}$, $8_\textrm{Bag}$, $9_\textrm{Ankle boot}\}$, $B$=$\{1, 3, 5, 7, 9\}$). 
For S2M sampling, samples are obtained at 100 MCMC iterations.

\textbf{Quantitative Results.}
Table~\ref{table:exp_MNIST} shows the results of our S2M sampling method and the baselines for the non-overlapping classes. The performance of GenPU is reported only for MNIST-3/5 due to its mode collapse issue~\citep{ChenLWZW20, ChiaroniKRHD20}. S2M sampling adopted to unconditional GAN shows promising results in terms of both accuracy and $\textrm{FID}^\dagger$. In fact, our results are superior to the existing methods specially designed for generating non-overlapping data such as GenPU and RumiLSGAN. Notably, S2M sampling improves accuracy by $8.5\%$ and $6.8\%$ while decreasing $\textrm{FID}^\dagger$ compared to the second-best models on MNIST-Even and FMNIST-Even, respectively. This indicates that S2M sampling accurately samples the images of the non-overlapping classes without being biased to a specific mode.

\input{Tables/Table_MNIST}

\textbf{Qualitative Results.}
Figure~\ref{fig:exp_MNIST} shows the qualitative results of RumiLSGAN, CPGAN and our method adopted to WGAN-GP. As indicated by the red boxes, RumiLSGAN and CP-GAN fail to completely eliminate the samples of the smaller class (\eg, odd digits for MNIST-Even or \emph{Sneaker} for FMNIST-Even). Conversely, S2M sampling correctly draws samples for the target classes.

\input{Figures/Figure_MNIST}


\subsection{Sampling for Multiple Classes}
\label{sec:exp_real_data}

\textbf{Experimental Settings.} 
Here, we consider a general case of single positive setting where multiple classes can be overlapping as shown in Figure~\ref{fig:exp_settings}~(b-d). 
Given $n$ classes, at most $2^n-1$ joint classes can be obtained. In this setting, we attempt to obtain samples of these joint classes using only single positive labels. We consider cGANs with a projection discriminator (cGAN-PD)~\citep{MiyatoK18}, AC-GAN~\citep{OdenaOS17}, and CP-GAN~\citep{KanekoUH19} as the baseline generative models.

Since traditional conditional GANs such as cGAN-PD and AC-GAN are not originally designed for generating joint classes, we naively introduce these models in our setting with slight modifications. Concretely, as a method to generate images belonging to $n$ classes, we provide $1/n$ as the conditional value for each class, following Kaneko \etal~\cite{KanekoUH19}. In our experiments, we expect these models to have a lower bound performance of our results and indicate their results with asterisks~($^\ast$).
In contrast, conditional GANs trained on a fully annotated dataset where all joint classes are specified can be considered as strong baselines in our settings. 
cGAN-PD and AC-GAN are trained in this setting, and are denoted as \emph{oracle models}.
In the experiments, S2M sampling is adopted on unconditional GAN and cGAN-PD. When adopting S2M sampling to cGAN-PD, a class containing the target joint class is used as the conditional value of cGAN-PD. BigGAN~\citep{BrockDS19} is used as the backbone architecture for all generative models, and only single positive labels are used except for the oracle models.

To demonstrate the effectiveness of our S2M sampling method, we use three real-world image datasets:
(i) \textbf{CIFAR-7to3} that has three classes ($A$, $B$, $C$), each of which contains four original classes of CIFAR-10, 
\ie $A$=\{\emph{Airplance}, \emph{Automobile}, \emph{Dog}, \emph{Frog}\}, $B$=\{\emph{Automobile}, \emph{Bird}, \emph{Cat}, \emph{Frog}\}, $C$=\{\emph{Cat}, \emph{Deer}, \emph{Dog}, \emph{Frog}\},
(ii) \textbf{CelebA-BMS} that consists of human face images, each of which is labeled for one of three attributes (\emph{Black-hair}, \emph{Male}, and \emph{Smiling}), and (iii) \textbf{CelebA-ABMNSW} that consists of human face images, each of which is labeled for one of six attributes (\emph{bAngs}, \emph{Black-hair}, \emph{Male}, \emph{No-beard}, \emph{Smiling}, and \emph{Wearing-lipstick}).
As depicted in Figure~\ref{fig:exp_settings}~(b-d), these datasets can have up to $2^3-1$ or $2^6-1$ joint classes.
For S2M sampling, samples are obtained at 200 MCMC iterations for CIFAR-7to3 and CelebA-BMS, and obtained at 1000 MCMC iterations for CelebA-ABMNSW.

\input{Figures/Figure_Real_Images_4x4}
\input{Tables/Table_Experiments_Real_Data}

\textbf{Quantitative Results.}
Table~\ref{table:exp_real_data} summarizes the quantitative results of joint class generation on CIFAR-7to3, CelebA-BMS, and CelebA-ABMNSW. As expected, we can observe that $\textrm{cGAN-PD}^\ast$ and $\textrm{AC-GAN}^\ast$ cannot accurately generate samples from joint classes, which is consistent with the findings of Kaneko~\etal~\cite{KanekoUH19}. 
Although CP-GAN shows reasonable accuracy for all experiments, it suffers from generating diverse samples. 
As discussed in Section~\ref{sec:introduction}, the high FID and low coverage support that CP-GAN tends to generate samples in a narrow scope of real data space.

In all experiments, our S2M sampling method adopted to cGAN-PD consistently outperformes the non-oracle baselines in terms of accuracy and diversity, \eg, in CIFAR-7to3, our method improves accuracy and FID of CP-GAN by $38\%$ and 13.74, respectively. Such improvements verify that our S2M sampling method generates correct samples for joint classes, and also preserves diversity within them.
Despite a large sample space of GAN, S2M sampling adopted to unconditional GAN also surpasses non-oracle baselines in terms of both fidelity and diversity while achieving reasonable accuracy. 
Surprisingly, despite the fact that the oracle models are trained with fully annotated data, S2M sampling shows a comparable performance against those models using only single positive labels.
Moreover, as an post-processing method, S2M sampling fully preserves the image quality of GANs and shows the highest fidelity in terms of precision and density.

\textbf{Qualitative Results.}
Figure~\ref{fig:exp_real_images} depicts the results of S2M sampling for every possible joint classes on CIFAR-7to3 and CelebA-BMS. The results show that S2M sampling can draw diverse high-quality images from the distributions of all joint classes.
Furthermore, S2M sampling is model-agnostic and does not require the retraining of GANs; thus, it can be easily integrated with various GANs including state-of-the-art GANs. We depict more qualitative results of adopting our S2M sampling method to unconditioanl GAN and pretrained StyleGANv2 in Appendix~\ref{appendix:more_results} and Appendix~\ref{appendix:diff_attributes}.

\input{Figures/Figure_GMM}

\textbf{Analysis on Latent Adaptation.}
To demonstrate the effect of latent adaptation, we tested our latent adaptation technique in a overlapping class (B+M+S) and a non-overlapping class (B-M-S) on the CelebA-BMS dataset. 
We first perform S2M sampling on unconditional BigGAN in these classes and draw $10k$ pilot latent samples obtained at 100 MCMC iterations. Using these latent samples, a Gaussian mixture of eight components is fit as a new latent of GAN (See Appendix~\ref{appendix:s2m_latent_adaptation}).
Due to the discrepancy between the generator distribution and the target distribution, S2M sampling originally shows a low acceptance probability, as shown in Figure~\ref{fig:exp_gmm}.
In this case, applying latent adaptation increases the acceptance probability and significantly improves both accuracy and FID at the early stage of iterations.
This indicates that the generator distribution becomes closer to the target distribution by adopting latent adaptation, resulting in an accelerated convergence speed.

%% file: Figures/Figure_Experiments_Settings.tex
\begin{figure*}[t!]
\begin{center}
    \includegraphics[width=1.0\linewidth]{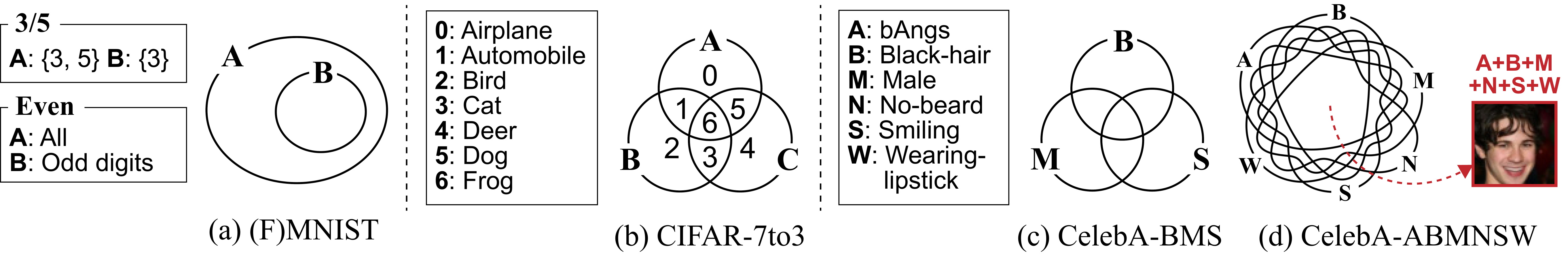}
\end{center}
\vspace{-0.15in}
    \caption{Experimental settings for (F)MNIST, CIFAR-7to3, CelebA-BMS, and CelebA-ABMNSW.}
\vspace{-0.25in}
\label{fig:exp_settings}
\end{figure*}

%% file: Tables/Table_MNIST.tex
\begin{table}[h!]
\begin{center}
\vspace{-0.1in}
\caption{Results for different models on MNIST and FMNIST.}
\label{table:exp_MNIST}
\scriptsize
\begin{tabular}[\linewidth]{l cc cc cc}
\toprule
\multirow{2}{*}{Method} &
\multicolumn{2}{c}{MNIST-3/5} &
\multicolumn{2}{c}{MNIST-Even} &
\multicolumn{2}{c}{FMNIST-Even} \\

\cmidrule(lr){2-3} \cmidrule(lr){4-5} \cmidrule(lr){6-7}
& Acc. ($\%$) ($\uparrow$) & $\textrm{FID}^\dagger$ ($\downarrow$)
& Acc. ($\%$) ($\uparrow$) & $\textrm{FID}^\dagger$ ($\downarrow$)
& Acc. ($\%$) ($\uparrow$) & $\textrm{FID}^\dagger$ ($\downarrow$) \\

\midrule

GenPU~\citep{HouCLZ18}
& \cell[99.33]{0.56} & \cell[1.93]{1.10} & - & - & - & - \\ 
RumiLSGAN~\citep{AsokanS20}
& \cell[77.06]{1.54} & \cell[13.20]{1.19} & \cell[86.11]{4.83} & \cell[3.44]{1.39} & \cell[91.07]{0.88} & \cell[3.23]{2.48} \\ 
CP-GAN~\citep{KanekoUH19}
& \cell[66.89]{1.46} & \cell[19.50]{0.90} & \cell[87.87]{0.40} & \cell[2.23]{0.08} & \cell[81.21]{0.67} & \cell[6.14]{0.56} \\
\textbf{GAN + Ours}
& \bcell[99.52]{0.34} & \bcell[0.88]{0.21} & \bcell[96.37]{0.25} & \bcell[0.86]{0.24} & \bcell[97.87]{0.66} & \bcell[2.48]{0.29} \\ 

\bottomrule
\end{tabular}
\end{center}
\vspace{-0.1in}
\end{table}

%% file: Figures/Figure_MNIST.tex
\begin{figure}[t!]
\begin{center}
    \includegraphics[width=1.0\linewidth]{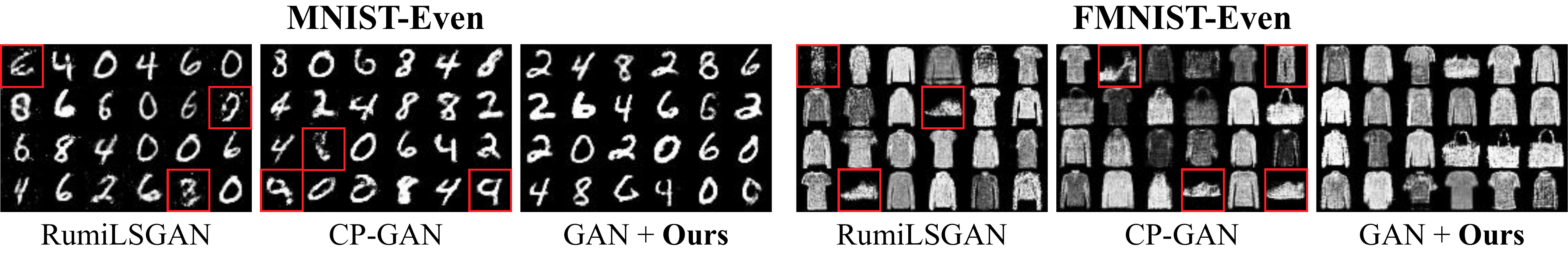}
\end{center}
\vspace{-0.15in}
    \caption{Qualitative results on MNIST-Even and FMNIST-Even.}
\vspace{-0.2in}
\label{fig:exp_MNIST}
\end{figure}

%% file: Figures/Figure_Real_Images_4x4.tex
\begin{figure*}[t!]
\begin{center}
    \includegraphics[width=1.0\linewidth]{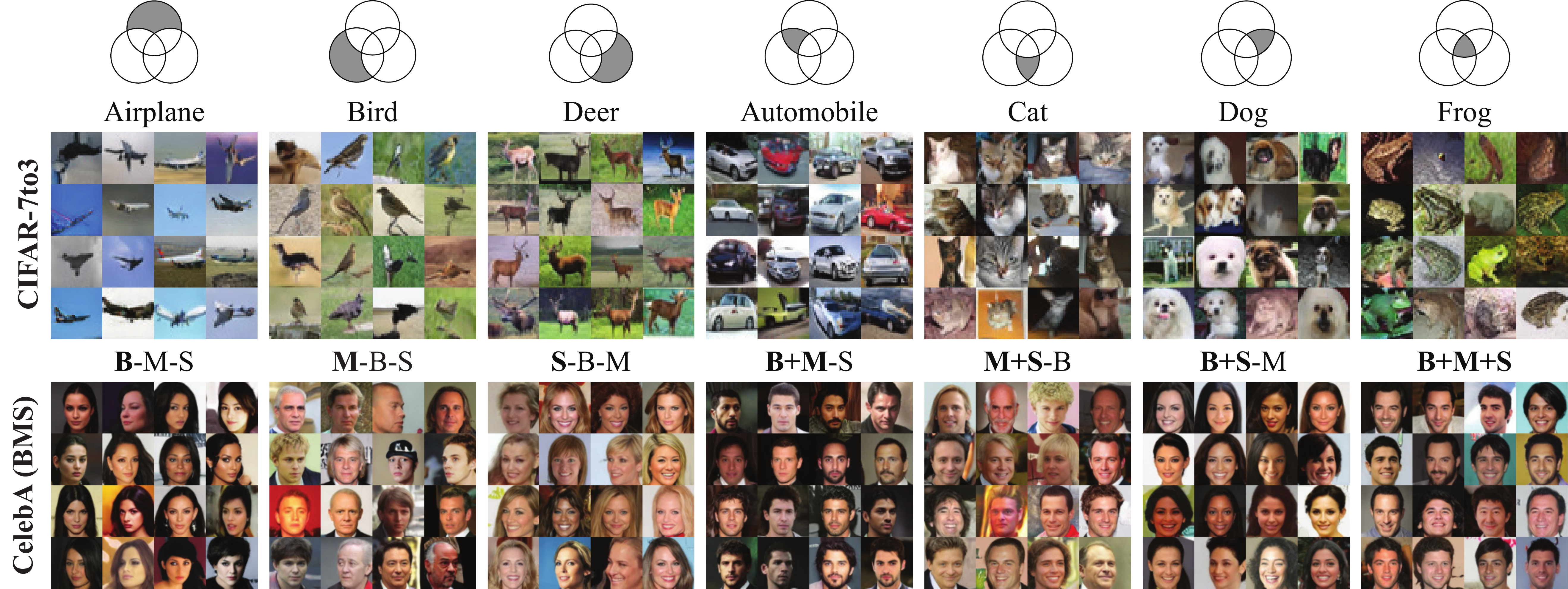}
\end{center}
\vspace{-0.1in}
    \caption{Results of S2M sampling with cGAN-PD on CIFAR-7to3 and CelebA-BMS. The first row depicts the target joint class.}
\label{fig:exp_real_images}
\end{figure*}

%% file: Tables/Table_Experiments_Real_Data.tex
\begin{table*}[t!]
\begin{center}
\vspace{-0.2in}
\caption{Accuracy~(\%), FID, IS, precision~(P), recall~(R), density~(D), and coverage~(C) for different models on real-world datasets.}
\label{table:exp_real_data}
\begin{adjustbox}{width=1\textwidth}
\scriptsize
\begin{tabular}[1.0\linewidth]{l l ccccccc}
\toprule
& Model & Acc. ($\uparrow$) & FID ($\downarrow$) & IS ($\uparrow$)  &  P ($\uparrow$) & R ($\uparrow$)  & D ($\uparrow$)  &  C ($\uparrow$) \\

\midrule\midrule

\multirow{7}{*}{\rotatebox[origin=c]{90}{CIFAR-7to3}}
& cGAN-PD (Oracle)~\citep{MiyatoK18} & \cell[87.60]{0.15} & \cell[16.55]{0.69} & \cell[8.39]{0.18} & \cell[0.73]{0.01} & \cell[0.65]{0.01} & \cell[0.89]{0.05} & \cell[0.85]{0.02} \\
& AC-GAN (Oracle)~\citep{OdenaOS17} & \cell[94.14]{0.39} & \cell[16.14]{0.43} & \cell[8.44]{0.03} & \cell[0.74]{0.00} & \cell[0.63]{0.00} & \cell[0.89]{0.02} & \cell[0.84]{0.00} \\
\cmidrule{2-9}
& $\text{cGAN-PD}^\ast$~\citep{MiyatoK18} & \cell[27.19]{0.53} & \cell[21.12]{0.62} & \cell[7.94]{0.02} & \cell[0.71]{0.01} & \lcell[0.66]{0.00} & \cell[0.81]{0.03} & \cell[0.78]{0.01} \\
& $\text{AC-GAN}^\ast$~\citep{OdenaOS17} & \cell[31.06]{1.17} & \cell[26.42]{0.82} & \cell[7.31]{0.10} & \cell[0.69]{0.01} & \lcell[0.66]{0.01} & \cell[0.71]{0.01} & \cell[0.70]{0.01} \\
& CP-GAN~\citep{KanekoUH19} & \cell[42.62]{0.54} & \cell[27.88]{2.00} & \cell[7.42]{0.06} & \cell[0.71]{0.00} & \lcell[0.66]{0.00} & \cell[0.76]{0.00} & \cell[0.71]{0.02} \\
& \textbf{GAN + Ours} & \lcell[77.65]{1.22} & \lcell[14.42]{0.55} & \lcell[8.71]{0.16} & \lcell[0.72]{0.00} & \textbf{\cell[0.67]{0.00}} & \lcell[0.92]{0.02} & \lcell[0.87]{0.01} \\
& \textbf{cGAN-PD + Ours} & \bcell[80.62]{2.08} & \bcell[14.14]{0.34} & \bcell[9.06]{0.18} & \bcell[0.75]{0.01} & \lcell[0.66]{0.01} & \bcell[1.09]{0.12} & \bcell[0.90]{0.01} \\
\midrule\midrule

\multirow{7}{*}{\rotatebox[origin=c]{90}{CelebA-BMS}}
& cGAN-PD (Oracle)~\citep{MiyatoK18} & \cell[78.35]{1.99} & \cell[10.38]{0.55} & \cell[2.45]{0.07} & \cell[0.79]{0.03} & \cell[0.48]{0.02} & \cell[1.12]{0.18} & \cell[0.86]{0.02} \\
& AC-GAN (Oracle)~\citep{OdenaOS17} & \cell[91.57]{0.43} & \cell[10.41]{1.52} & \cell[2.40]{0.03} & \cell[0.79]{0.02} & \cell[0.49]{0.04} & \cell[1.18]{0.12} & \cell[0.86]{0.03} \\
\cmidrule{2-9}
& $\text{cGAN-PD}^\ast$~\citep{MiyatoK18} & \cell[39.84]{0.70} & \cell[10.97]{1.56} & \cell[2.32]{0.06} & \cell[0.79]{0.01} & \lcell[0.48]{0.04} & \cell[1.19]{0.05} & \cell[0.86]{0.02} \\
& $\text{AC-GAN}^\ast$~\citep{OdenaOS17} & \cell[52.01]{1.81} & \cell[12.39]{1.02} & \cell[2.32]{0.10} & \cell[0.79]{0.02} & \textbf{\cell[0.49]{0.01}} & \cell[1.17]{0.09} & \cell[0.85]{0.02} \\
& CP-GAN~\citep{KanekoUH19} & \lcell[87.36]{2.05} & \cell[13.38]{1.36} & \cell[2.43]{0.09} & \cell[0.78]{0.01} & \cell[0.45]{0.02} & \cell[1.08]{0.04} & \cell[0.82]{0.02} \\
& \textbf{GAN + Ours} & \cell[85.22]{4.27} & \textbf{\cell[10.50]{0.97}} & \textbf{\cell[2.51]{0.08}} & \lcell[0.83]{0.02} & \lcell[0.48]{0.03} & \lcell[1.36]{0.08} & \textbf{\cell[0.89]{0.02}} \\
& \textbf{cGAN-PD + Ours} & \textbf{\cell[90.44]{1.05}} & \lcell[10.63]{0.29} & \lcell[2.46]{0.06} & \textbf{\cell[0.84]{0.00}} & \lcell[0.48]{0.02} & \textbf{\cell[1.40]{0.02}} & \lcell[0.88]{0.01} \\
\midrule\midrule

\multirow{7}{*}{\rotatebox[origin=c]{90}{CelebA-ABMNSW}}
& cGAN-PD (Oracle)~\citep{MiyatoK18} & \cell[69.61]{2.91} & \cell[13.54]{2.54} & \cell[2.44]{0.11} & \cell[0.79]{0.02} & \cell[0.42]{0.04} & \cell[1.15]{0.10} & \cell[0.84]{0.02} \\
& AC-GAN (Oracle)~\citep{OdenaOS17} & \cell[91.37]{1.14} & \cell[13.46]{1.29} & \cell[2.30]{0.05} & \cell[0.80]{0.02} & \cell[0.36]{0.01} & \cell[1.17]{0.09} & \cell[0.82]{0.01} \\
\cmidrule{2-9}
& $\text{cGAN-PD}^\ast$~\citep{MiyatoK18} & \cell[15.28]{0.08} & \cell[11.09]{0.67} & \cell[2.32]{0.03} & \cell[0.80]{0.01} & \textbf{\cell[0.47]{0.03}} & \cell[1.19]{0.07} & \cell[0.85]{0.02} \\
& $\text{AC-GAN}^\ast$~\citep{OdenaOS17} & \cell[23.97]{0.64} & \cell[11.92]{2.06} & \cell[2.44]{0.06} & \cell[0.77]{0.04} & \cell[0.45]{0.02} & \cell[1.11]{0.13} & \cell[0.82]{0.04} \\
& CP-GAN~\citep{KanekoUH19} & \lcell[79.09]{1.14} & \cell[14.97]{4.18} & \lcell[2.45]{0.04} & \cell[0.74]{0.03} & \lcell[0.46]{0.03} & \cell[0.95]{0.08} & \cell[0.77]{0.06} \\
& \textbf{GAN + Ours} & \cell[70.04]{2.97} & \bcell[9.77]{0.22} & \bcell[2.62]{0.05} & \lcell[0.81]{0.01} & \bcell[0.47]{0.01} & \lcell[1.28]{0.04} & \lcell[0.87]{0.01} \\
& \textbf{cGAN-PD + Ours} & \textbf{\cell[79.74]{0.78}} & \lcell[10.30]{0.48} & \lcell[2.45]{0.06} & \textbf{\cell[0.85]{0.01}} & \textbf{\cell[0.47]{0.02}} & \textbf{\cell[1.45]{0.06}} & \textbf{\cell[0.88]{0.00}} \\

\bottomrule
\end{tabular}
\end{adjustbox}
\end{center}
\vspace{-0.2in}
\end{table*}

%% file: Figures/Figure_GMM.tex
\begin{figure*}[t!]
\begin{center}
    \includegraphics[width=1.0\linewidth]{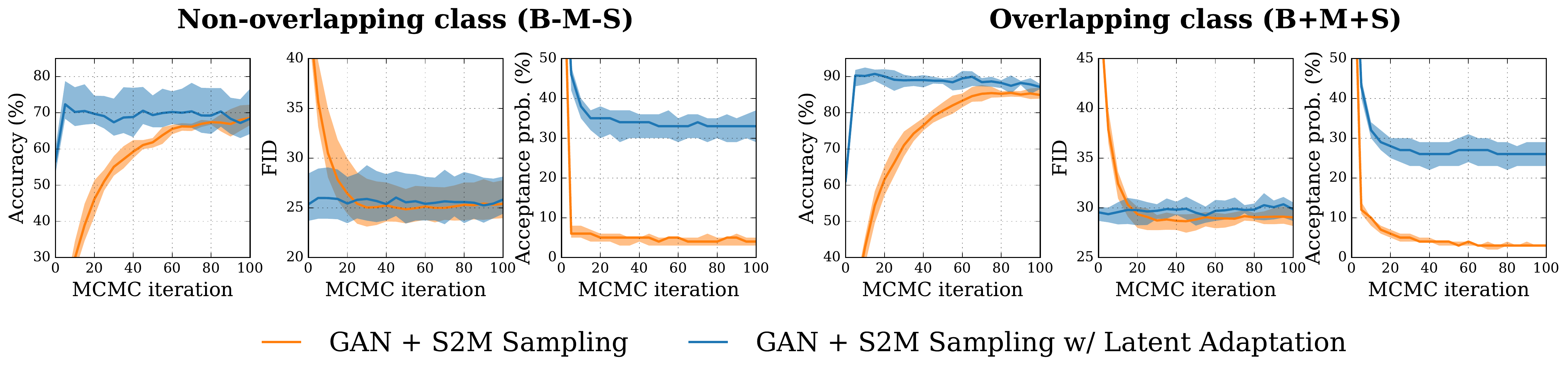}
\end{center}
\vspace{-0.2in}
    \caption{Accuracy (\%), FID, and acceptance probability (\%) per MCMC iteration for S2M sampling with and without latent adaptation.
    }
\vspace{-0.2in}
\label{fig:exp_gmm}
\end{figure*}

%% file: 05_Limitations.tex
\section{Limitations}
\label{sec:limitations}

\input{Figures/Figure_CelebA_Statics}

In general, as the number of attributes increases, the number of each joint class samples in the training dataset are drastically reduced. This may cause two primary problems: (i) training GANs to be able to generate each joint class samples becomes challenging, and S2M sampling cannot draw samples in which GANs do not generate, and (ii) the computation time of the sampling procedure increases. In this case, it may be necessary to apply the latent adaptation technique to shorten MCMC chains. 
In principle, we can reduce the sampling time complexity to be proportional to the number of attributes (See Appendix~\ref{appendix:time_complexity}).

As shown in Figure~\ref{fig:celeba_stat}, the number of training samples in joint classes rapidly decreases. Because of this, six attributes at most are used on the CelebA dataset for experiments. Almost half of the joint classes have fewer than 100 training samples when using six attributes (\ie, CelebA-ABMNSW). 
Nevertheless, our method handles a larger number of attributes compared to the existing methods. We believe that our work can be further extended to larger attributes with the sufficient amount of data.

%% file: Figures/Figure_CelebA_Statics.tex
\begin{wrapfigure}{r}{0.46\textwidth}
\vspace{-0.2in}
    \includegraphics[width=1.0\linewidth]{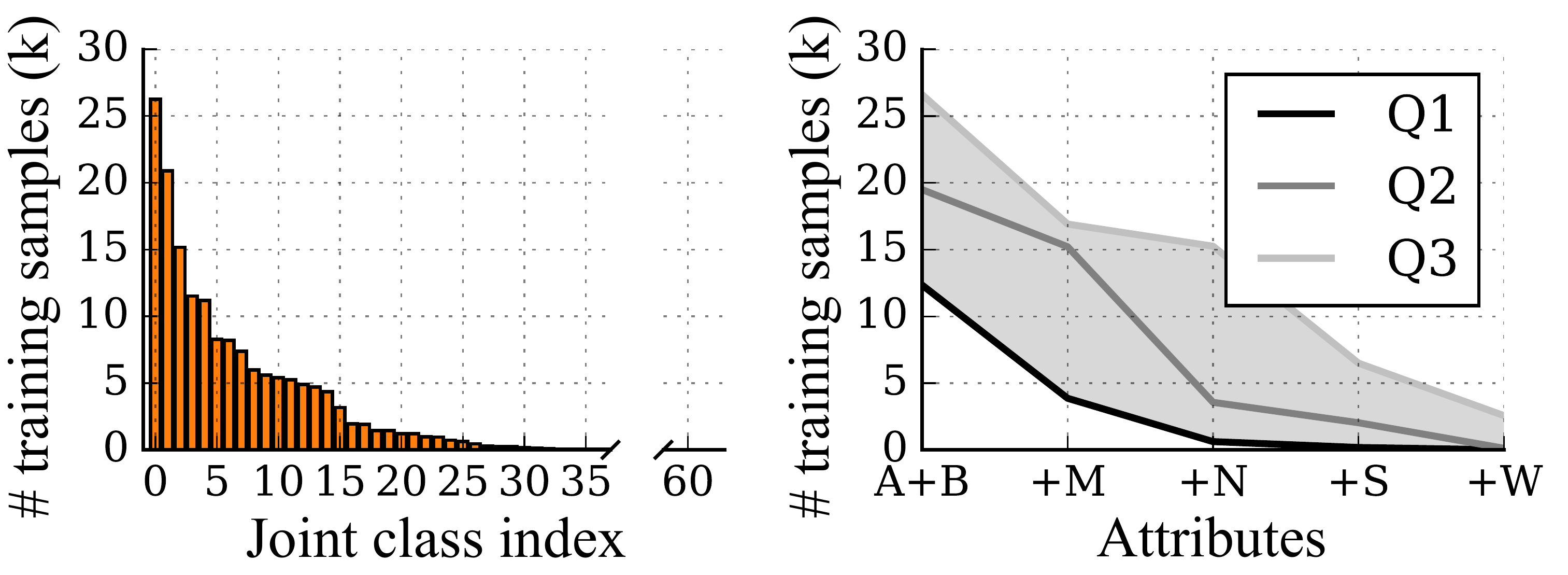}
\vspace{-0.25in}
    \caption{(Left) Distribution of the number of training samples for each joint class on CelebA-ABMNSW. (Right) As attributes are added, the quartiles of the number of samples for each joint class significantly decrease.
}
\vspace{-0.15in}
\label{fig:celeba_stat}
\end{wrapfigure}

%% file: 06_Conclusion.tex
\section{Conclusion}

In this study, we investigate the single positive setting in the class-conditional generation task and propose a novel sampling framework called S2M sampling.
We demonstrate that our proposed S2M sampling method can accurately draw high quality multi-label samples in the single positive setting. To improve sampling efficiency, we also introduce the latent adaptation technique.
We believe that our theoretical framework provides a better understanding for weakly-supervised multi-label data generation.
Our S2M sampling method has the potential to be applied to a wide variety of generative models as well, including clustering based GANs~\cite{CasanovaCVDR21} and diffusion models~\cite{HoJA20, DhariwalN21}, among others.

%% file: Appendix.tex
\newpage
\appendix

\input{Appendix/A_Proof}
\input{Appendix/B_Algorithm}
\input{Appendix/C_Time_Complexity}
\input{Appendix/D_Experiments_Details}
\input{Appendix/E_Experimental_Results}
\input{Appendix/F_Varying_Labels}
\input{Appendix/G_Consideration}
\input{Appendix/Social_Impacts}

%% file: Appendix/A_Proof.tex
\section{Proof of Main Theorem}
\label{appendix:proof}

Let $\{f_{(I,J)}:X\to[0,\infty)\}_{(I,J)\in\indexset}$ be an indexed family of non-negative measurable functions on $X$, and let $f_k\coloneqq f_{(\{k\},\emptyset)}$ for $k\in N$.
We will consider two following properties:
\begin{enumerate}[label=(\alph*)]
\item $\forall (I, J)\in \indexset, f_{(I, J)} = \sum_{S:I\subseteq S, J\subseteq N\setminus S} f_{(S,N\setminus S)}$
\item $\forall i, j\in N \text{ s.t. } i\neq j, \supp f_{(\{i\},\{j\})} \cap \supp f_{(\{j\},\{i\})} = \emptyset$
\end{enumerate}
We first derive some useful lemmas to prove the necessity of Theorem~\ref{thm:equivalence}.

\begin{lem}
\label{lem:disjoint}
Assume that (a) and (b) hold. $\forall (I_1, J_1), (I_2, J_2)\in \indexset$ s.t. $I_1 \cap J_2 \neq \emptyset, J_1 \cap I_2 \neq \emptyset$, $\supp f_{(I_1,J_1)} \cap \supp f_{(I_2,J_2)} = \emptyset$.
\end{lem}
\begin{proof}
Choose any $u\in I_1 \cap J_2, v\in J_1 \cap I_2$. By (a), 
\begin{equation}
\begin{aligned}
f_{(\{u\},\{v\})} & = \sum\nolimits_{S:u\in S, v\in N\setminus S} f_{(S,N\setminus S)} \ge \sum\nolimits_{S:I_1\subseteq S, J_1\subseteq N\setminus S} f_{(S,N\setminus S)} = f_{(I_1,J_1)}\\
f_{(\{v\},\{u\})} & = \sum\nolimits_{S:v\in S, u\in N\setminus S} f_{(S,N\setminus S)} \ge \sum\nolimits_{S:I_2 \subseteq S, J_2 \subseteq N\setminus S} f_{(S,N\setminus S)} = f_{(I_2,J_2)}.
\end{aligned}
\end{equation}
If $u=v$, then $I_1 \cap J_1 \neq \emptyset$ which contradicts the definition of $\indexset$. Therefore, $u\neq v$. By (b), $\supp f_{(I_1,J_1)} \cap \supp f_{(I_2,J_2)} \subseteq \supp f_{(\{u\},\{v\})} \cap \supp f_{(\{v\},\{u\})} = \emptyset$.
\end{proof}

\begin{lem}
\label{lem:disjoint_sum}
Assume that (a) and (b) hold. $\forall i \in N, \forall I \subseteq N$ s.t. $\{i\} \subsetneq I$, $\supp f_{(I\setminus\{i\},\{i\})} \cap \supp \sum\nolimits_{S:i\in S, I\nsubseteq S \subseteq N} f_{(S,N\setminus S)} = \emptyset$.
\end{lem}
\begin{proof}
Let $S\subseteq N$ s.t. $i\in S \nsupseteq I$. Then, $\emptyset \neq I\setminus S \subseteq \left(I\setminus\{i\}\right) \cap \left(N\setminus S\right)$. By Lemma~\ref{lem:disjoint}, $\supp f_{(I\setminus\{i\},\{i\})} \cap \supp f_{(S,N\setminus S)} = \emptyset$. Therefore, $\supp f_{(I\setminus\{i\},\{i\})} \cap \supp \sum\nolimits_{S:i\in S, I\nsubseteq S \subseteq N} f_{(S,N\setminus S)} = \emptyset$.
\end{proof}

\begin{lem}
\label{lem:min}
Assume that (a) and (b) hold. $\forall I\subseteq N$ s.t. $I\neq \emptyset$, $f_{(I, \emptyset)}=\min_{i\in I}f_i$.
\end{lem}
\begin{proof}
We will use induction to prove the lemma. Let $P(k)$ be the following statement.
\begin{equation}
P(k): \forall I \text{ s.t. } 1\leq|I|=k\leq |N|, \text{then } f_{(I, \emptyset)}=\min_{i\in I}f_i.
\end{equation}
For the base case $k=1$, the statement holds by the definition. Assume that the induction hypothesis for $k\leq l<|N|$ holds. Consider $|I|=l+1$ and choose any $i\in I$. Then,
\begin{equation}
\begin{aligned}
\min_{i\in I}f_i & = \min\{f_{(I\setminus\{i\},\emptyset)}, f_{(\{i\},\emptyset)}\} && \text{by the induction hypothesis}\\
& = f_{(I\setminus\{i\},\emptyset)} - \max\{f_{(I\setminus\{i\},\emptyset)}-f_{(\{i\},\emptyset)}, 0\} \\
& = f_{(I\setminus\{i\},\emptyset)} - \max\{f_{(I\setminus\{i\},\{i\})}-\sum\nolimits_{S:i\in S, I\nsubseteq S \subseteq N} f_{(S,N\setminus S)}, 0\} \\
& = f_{(I\setminus \{i\},\emptyset)} - f_{(I\setminus \{i\},\{i\})} &&\text{by Lemma~\ref{lem:disjoint_sum}}\\
& = f_{(I,\emptyset)}
\end{aligned}
\end{equation}
Therefore, $P(l+1)$ holds. We conclude that $f_{(I, \emptyset)}=\min_{i\in I}f_i$ for $\emptyset \neq I\subseteq N$.
\end{proof}

\universe*

\begin{proof}
We first show the necessity of the condition. Assume that (a) and (b) hold. If $J=\emptyset$, then $f_{(I, J)}=\min_{i\in I}f_i$ by Lemma2. Hence, we may assume that $J\neq \emptyset$. Fix $x\in X$, and let $\{a_1, a_2, ..., a_{|J|}\}$ be an arrangement of $J$ so that $f_{a_i}(x) \leq f_{a_j}(x)$ for all $i<j$. For every $\emptyset \neq S\subseteq J$, we let $m(S)$ denote the minimum index $s$ such that $a_s\in S$.

Note that
\begin{equation}
\begin{aligned}
    f_{(I, J)}(x) & = \sum_{S\subseteq J} (-1)^{|S|} f_{(I\cup S, \emptyset)}(x) && \text{by Inclusion–exclusion principle} \\
    & = \sum_{S\subseteq J} (-1)^{|S|} \min_{i\in I\cup S}f_i(x) && \text{by Lemma~\ref{lem:min}} \\
\end{aligned}
\end{equation}
We now decompose the last summation into three cases.
\begin{enumerate}[label=(\roman*)]
\item $S=\emptyset$
\begin{equation}
(-1)^{|S|}\min_{i\in I\cup S}f_i(x)=\min_{i\in I}f_i(x).
\end{equation}
\item $m(S)<|J|$
\begin{equation}
\begin{aligned}
    \sum_{S:m(S)<|J|} (-1)^{|S|} \min_{i \in I\cup S}f_i(x)
    & = \sum_{j<|J|}\sum_{S:m(S)=j} (-1)^{|S|} \min_{i\in I\cup \{a_j\}}f_i(x) \\
    & = \sum_{j<|J|} \left(\min_{i\in I\cup \{a_j\}}f_i(x)\right) \left\{ (-1)\cdot 2^{|J|-j-1}+2^{|J|-j-1}\right\} \\
    & = 0.
\end{aligned}
\end{equation}
\item $m(S)=|J|$
\begin{equation}
(-1)^{|S|}\min_{i\in I\cup S}f_i(x)=-\min_{i\in I\cup\{a_{|J|}\}}f_i(x).
\end{equation}
\end{enumerate}
Summing up all of the above terms gives the rest result.
\begin{equation}
\begin{aligned}
    f_{(I, J)}(x) & = \min_{i\in I}f_i(x) -\min_{i\in I\cup\{a_{|J|}\}}f_i(x) \\
    & = \min_{i\in I}f_i(x) - \min\{\min_{i\in I}f_i(x), \max_{j\in J}{f_j(x)}\} \\
    & = \left(\min_{i\in I}f_i(x) - \max_{j\in J}f_j(x)\right)^+.
\end{aligned}
\end{equation}
To show the sufficiency, assume
\begin{equation}
\forall (I, J) \in \indexset, f_{(I, J)}=
    \begin{cases}
    \left(\min_{i\in I}f_i - \max_{j\in J}f_j\right)^+ &\mbox{if } J\neq \emptyset \\
    \min_{i\in I}f_i & \mbox{otherwise}
    \end{cases}.
\end{equation}
Let us assume that $f_{(I, J)} \neq \sum_{S:I\subseteq S, J\subseteq N\setminus S} f_{(S,N\setminus S)}$ for some $(I, J)\in \indexset$. Choose such $I, J$ so that $|I|+|J|$ is maximum. Note that $I\cup J\subsetneq N$ because $\sum_{S:I\subseteq S, J\subseteq N\setminus S} f_{(S,N\setminus S)}$ is exactly the same expression as $f_{(I, J)}$ for $I\cup J=N$. Hence, we can choose some $k\in N\setminus (I\cup J)$. By the maximality
of $|I|+|J|$, the following two equations hold.
\begin{equation}
\begin{aligned}
f_{(I, J\cup \{k\})} & = \sum\nolimits_{S:I\subseteq S, J\cup \{k\}\subseteq N\setminus S} f_{(S,N\setminus S)} \\
f_{(I\cup \{k\}, J)} & = \sum\nolimits_{S:I\cup \{k\}\subseteq S, J\subseteq N\setminus S} f_{(S,N\setminus S)}.
\end{aligned}
\end{equation}

We use above two equations and consider all possible inequalities among $\min_{i\in I} f_i$, $max_{j\in J} f_j$, and $f_k$. The following equation always holds regardless of these inequalities.
\begin{equation}
\begin{aligned}
\sum_{S:I\subseteq S, J\subseteq N\setminus S} & f_{(S,N\setminus S)} = f_{(I, J\cup \{k\})} + f_{(I\cup \{k\}, J)} \\
& = 
\begin{cases}
    \left(\min_{i\in I}f_i - \max_{j\in J\cup \{k\}}f_j\right)^+ + \left(\min_{i\in I\cup \{k\}}f_i - \max_{j\in J}f_j\right)^+ &\mbox{if } J\neq \emptyset \\
    (\min_{i\in I}f_i - f_k)^+ + \min_{i\in I\cup \{k\}}f_i & \mbox{otherwise}
\end{cases}\\
& = 
\begin{cases}
    \left(\min_{i\in I}f_i - \max_{j\in J}f_j\right)^+ &\mbox{if } J\neq \emptyset \\
    \min_{i\in I}f_i & \mbox{otherwise}
\end{cases}\\
& = f_{(I, J)},
\end{aligned}
\end{equation}
which leads to a contradiction.

Also, for every $i, j \in N$ such that $i\neq j$,
\begin{equation}
    \begin{aligned}
    \min(f_{(\{i\},\{j\})},f_{(\{j\},\{i\})}) & = \min\{(f_i-f_j)^+,(f_j-f_i)^+\} \\
    & = \left(\min\{f_i-f_j,f_j-f_i\}\right)^+ \\
    & = 0.
    \end{aligned}
\end{equation}
Therefore, $(a)$ and $(b)$ hold.
\end{proof}

%% file: Appendix/B_Algorithm.tex
\section{Description for S2M Sampling}
\label{appendix:s2m_description}

\subsection{Pseudocode of S2M Sampling}
\label{appendix:s2m_pseudocode}
In Section~\ref{sec:s2m_sampling}, we describe how to build S2M sampling upon unconditional GANs and class-conditional GANs. Algorithm~\ref{alg:S2M_sampling} illustrates the use of S2M sampling for GANs. This algorithm can be easily modified to the conditional versions by replacing the acceptance probability $\alpha$ (See Equation~\ref{eq:alpha_c}).

\input{Algorithms/Alg_S2M_Sampling}

\subsection{Latent Adaptation with Gaussian Mixture Model}
\label{appendix:s2m_latent_adaptation}
In Section~\ref{sec:s2m_sampling}, we describe the latent adaptation technique to improve the sampling efficiency of our proposed S2M sampling method. In this section, we provide an example for the latent adaptation technique using a Gaussian mixture model.
The real examples of using latent adaption are shown in Figure~\ref{fig:la_example}.
After obtaining target latent samples $t_{1:m}$ from S2M sampling, we use those samples to fit a multivariate Gaussian mixture model $\tilde{p}_z(z)=\sum_{i=1}^M \phi_i \mathcal{N}(z|\mu_i, \Sigma_i)$. The parameters can be updated using an expectation–maximization algorithm~\footnote{Christopher M. Bishop. Pattern recognition and machine learning, 5th Edition. 2007.}
As explained in Section~\ref{sec:s2m_sampling}, we run the MH algorithm where the proposal $x' \sim q(x'|x)=\tilde{p}_G(x')$ is accepted with a probability $\tilde{\alpha}(x', x)$ which is calculated as
\begin{equation}
\begin{aligned}
\tilde{\alpha}(x', x) = \min\left(1, \frac{p_{(I,J)}(x')/\tilde{p}_G(x')}{p_{(I,J)}(x)/\tilde{p}_G(x)}\right) \approx \min\left(1,\frac{r_{(I,J)}(x')({D_v^\ast(x)}^{-1}-1)p_z(z')/\tilde{p}_z(z')}{r_{(I,J)}(x)({D_v^\ast(x')}^{-1}-1)p_z(z)/\tilde{p}_z(z)}\right).
\end{aligned}
\end{equation}
If the latent prior of the generator $G$ is the standard multivariate normal distribution $p_z(z) = \mathcal{N}(z|0, I_d)$, we can compute the acceptance probability by explicitly calculating the density ratio $p_z(z) / \tilde{p}_z(z)$ as follows:
\begin{equation}
\begin{aligned}
\frac{p_z(z)}{\tilde{p}_z(z)} = \frac{\mathcal{N}(z|\mu_0, \Sigma_0)}{\sum_{i=1}^M \phi_i \mathcal{N}(z|\mu_i, \Sigma_i)} = \left( \sum_{i=1}^M \phi_i |\Sigma_i|^{-\frac{1}{2}} \exp\left(-\frac{1}{2}(z-\mu_i)^T{\Sigma_i}^{-1} (z-\mu_i)+ \frac{1}{2}z^T z\right) \right)^{-1}.
\end{aligned}
\end{equation}
In our experiments, some determinants of the covariance matrices often approach zero, resulting in numerical errors. One way to readily avoid this problem is to let $\tilde{p}_z$ be a Gaussian mixture with a shared covariance matrix $\Sigma$, \ie, $\Sigma_1 = \Sigma_2 = ... = \Sigma_M=\Sigma$. Then, we no longer need to compute the determinants of the covariance matrices to calculate the acceptance probability because  $p_z(z')\tilde{p}_z(z)/\tilde{p}_z(z')/p_z(z)$ is simplified as follows:
\begin{equation}
\begin{aligned}
\frac{p_z(z') \tilde{p}_z(z)}{\tilde{p}_z(z') p_z(z)} = \frac{\sum_{i=1}^M \phi_i \exp\left(-\frac{1}{2}(z-\mu_i)^T{\Sigma}^{-1} (z-\mu_i)+ \frac{1}{2}z^T z\right)}{\sum_{i=1}^M \phi_i \exp\left(-\frac{1}{2}(z'-\mu_i)^T{\Sigma}^{-1} (z'-\mu_i)+ \frac{1}{2}{z'}^T z'\right)}.
\end{aligned}
\end{equation}
Algorithm~\ref{alg:latent_adaptation_sharing} illustrates the sampling process using the Gaussian mixture latent with a shared covariance matrix. After collecting the filtered latent samples $z_{1:s}$ through Algorithm~\ref{alg:latent_adaptation_sharing}, we can get target class samples $x_{1:s}$ by taking $x_{i}=G(z_i)$ for $i=1,2,...,s$.
As explained in Section~\ref{sec:s2m_sampling}, the latent adaptation technique can be applied iteratively by increasing the number of specified attributes one by one. Algorithm~\ref{alg:latent_adaptation_sharing_repeat} illustrates this strategy and Section~\ref{appendix:time_complexity} provides the computation complexity analysis of the algorithm.

\input{Algorithms/Alg_Latent_Adaptation_Sharing}
\input{Algorithms/Alg_Latent_Adaptation_Sharing_Repeat}

\input{Figures/Figure_LA_Example}

%% file: Algorithms/Alg_S2M_Sampling.tex
\begin{algorithm}[H]
	\caption{\emph{S2M Sampling for GANs}}
	\textbf{Input:} generator $G$, classifiers $D_v^\ast, D_r^\ast$, intersection index set $I$, difference index set $J$, and class prior ratios $\gamma_{1:N}$\\
    \textbf{Output:} filtered sample $x$
	\begin{algorithmic}[1]
    \State Choose any $x \in \supp p_{(I, J)}$.
	\For {$k=1$ to $K$}
	    \State Draw $x'$ from $G$.
	    \State Draw $u$ from Uniform(0,1).
	    \State $r_i \leftarrow \gamma_i D_r^\ast(i|x)$ for every $i\in I\cup J$
	    \State $r'_i \leftarrow \gamma_i D_r^\ast(i|x')$ for every $i\in I\cup J$
	    \State $\alpha \leftarrow \min\left(1,\frac{\left(\min\{r'_i: i\in I\} - \max\{r'_j:j\in J\}\cup\{0\}\right)^+({D_v^\ast(x)}^{-1}-1)}{\left(\min\{r_i: i\in I\} - \max\{r_j:j\in J\}\cup\{0\}\right)^+({D_v^\ast(x')}^{-1}-1)}\right)$
        \If {$u \leq \alpha$}
            \State $x \leftarrow x'$
        \EndIf
    \EndFor
\end{algorithmic}
\label{alg:S2M_sampling}
\end{algorithm}

%% file: Algorithms/Alg_Latent_Adaptation_Sharing.tex
\begin{algorithm}[H]
	\caption{\emph{Sampling with Adapted Latent Space}}
	\textbf{Input:} generator $G$, classifiers $D_v^\ast, D_r^\ast$, intersection index set $I$, difference index set $J$, class prior ratios $\gamma_{1:N}$ and target latent distribution parameters $\phi_{1:M}, \mu_{1:M}, \Sigma$\\
    \textbf{Output:} filtered latent sample $z$
	\begin{algorithmic}[1]
	\State Let $\tilde{p}_z(z)\coloneqq\sum_{i=1}^M \phi_i \mathcal{N}(z|\mu_i, \Sigma)$.
    \State Choose any $z$ such that $G(z) \in \supp p_{(I, J)}$.
	\For {$k=1$ to $K$}
	    \State Draw $z'$ from $\tilde{p}_z$.
	    \State Draw $u$ from Uniform(0,1).
	    \State $r_i \leftarrow \gamma_i D_r^\ast(i|G(z))$ for every $i\in I\cup J$
	    \State $r'_i \leftarrow \gamma_i D_r^\ast(i|G(z'))$ for every $i\in I\cup J$
	    \State $d \leftarrow \frac{\sum_{i=1}^M \phi_i \exp\left(-\frac{1}{2}(z-\mu_i)^T{\Sigma}^{-1} (z-\mu_i)+ \frac{1}{2}z^T z\right)}{\sum_{i=1}^M \phi_i \exp\left(-\frac{1}{2}(z'-\mu_i)^T{\Sigma}^{-1} (z'-\mu_i)+ \frac{1}{2}{z'}^T z'\right)}$
	    \State $\alpha \leftarrow \min\left(1,\frac{\left(\min\{r'_i: i\in I\} - \max\{r'_j:j\in J\}\cup\{0\}\right)^+({D_v^\ast(G(z))}^{-1}-1)}{\left(\min\{r_i: i\in I\} - \max\{r_j:j\in J\}\cup\{0\}\right)^+({D_v^\ast(G(z'))}^{-1}-1)} \cdot d\right)$
        \If {$u \leq \alpha$}
            \State $z \leftarrow z'$
        \EndIf
    \EndFor
\end{algorithmic}
\label{alg:latent_adaptation_sharing}
\end{algorithm}

%% file: Algorithms/Alg_Latent_Adaptation_Sharing_Repeat.tex
\begin{algorithm}[H]
	\caption{\emph{Repeating Latent Adaptation}}
	\textbf{Input:} generator $G$, classifiers $D_v^\ast, D_r^\ast$, intersection index set $I$, difference index set $J$, and class prior ratios $\gamma_{1:N}$\\
    \textbf{Output:} filtered latent samples $z_{1:s}$
	\begin{algorithmic}[1]
	\State Let $I_0 \subseteq I$ such that $\left|I_0\right|=1$, and let $J_0=\emptyset$.
	\State $t_{1:m}\leftarrow \text{MHAlgorithm}(G, D_v^\ast, D_r^\ast, I_0, J_0, \gamma_{1:N})$ \Comment{Obtain latent samples using Algorihtm~\ref{alg:S2M_sampling}}
    \While {$I_0 \cup J_0 \neq I \cup J$}
        \State Fit $\tilde{p}_z(z)=\sum_{i=1}^M \phi_i \mathcal{N}(z|\mu_i, \Sigma)$ with $t_{1:m}$ using the expectation-maximization algorithm.
    	\State Choose $\alpha \subseteq I\setminus I_0, \beta \subseteq J\setminus J_0$ such that $\left|\alpha\cup\beta\right|=1$.
        \State $I_0 \leftarrow I_0 \cup \{\alpha\}$
        \State $J_0 \leftarrow J_0 \cup \{\beta\}$
    	\State $t_{1:m}\leftarrow \text{AdaptiveLatentSampling}(G, D_v^\ast, D_r^\ast, I_0, J_0, \gamma_{1:N}, \phi_{1:M}, \mu_{1:M}, \Sigma)$
    	\Comment{Obtain latent samples using Algorihtm~\ref{alg:latent_adaptation_sharing}}
    \EndWhile
    \State Fit $\tilde{p}_z(z)=\sum_{i=1}^M \phi_i \mathcal{N}(z|\mu_i, \Sigma)$ with $t_{1:m}$ using the expectation-maximization algorithm.
    \State $z_{1:s}\leftarrow \text{AdaptiveLatentSampling}(G, D_v^\ast, D_r^\ast, I_0, J_0, \gamma_{1:N}, \phi_{1:M}, \mu_{1:M}, \Sigma)$
    \Comment{Obtain latent samples using Algorihtm~\ref{alg:latent_adaptation_sharing}}
\end{algorithmic} 
\label{alg:latent_adaptation_sharing_repeat}
\end{algorithm}

%% file: Figures/Figure_LA_Example.tex
\begin{figure}[h]
\begin{center}
    \includegraphics[width=1.0\linewidth]{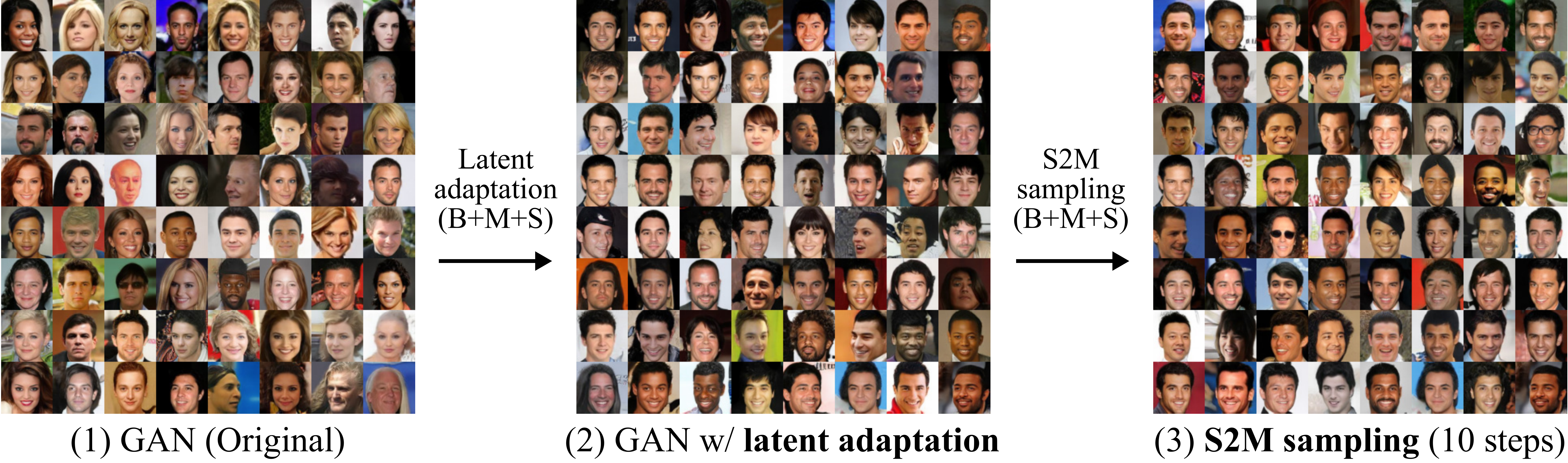}
\end{center}
\caption{Results of sequentially applying latent adaptation and S2M sampling for \emph{B+M+S} joint class. (1) Before applying latent adaptation, GANs draw diverse CelebA images. (2) After applying latent adaptation to B+M+S class, the generator produces images of that class with a high probability. (3) By using latent adaptation, S2M sampling can remove samples outside of that class even with a few MCMC iterations.}
\label{fig:la_example}
\end{figure}

%% file: Appendix/C_Time_Complexity.tex
\section{Computational Complexity Analysis}
\label{appendix:time_complexity}

\textbf{Inference time of different architectures.} In this section, we compare the inference time of several architectures as shown in Table~\ref{table:running_time}. The classifiers and Gaussian mixture sampler add a small amount of time to the inference process for S2M sampling. To account for these components, we measure the inference time per iteration during the sampling procedure. We use a single RTX 3090 GPU for this experiment.

\input{Tables/Table_running_time}

\textbf{Time complexity of sampling algorithms.} To simplify the analysis, we assume that a generator distribution is close to a real distribution, and let $\alpha$ be the probability of a generated sample belonging to a target joint class. 
Then, we have $\frac{p_{(I,J)}(x)}{p_G(x)} \leq \frac{1}{\alpha}$ for all $x$. By Theorem 2.1 of Mengersen \etal~\cite{1033066201} 
the convergence speed of the independent MH algorithm in this condition is given by
\begin{equation}
||P^{t}(x,\cdot)-p_{(I,J)} ||_{TV} \leq (1-\alpha)^t,
\end{equation}
where $||\cdot||_{TV}$ is the total variation distance and $P^t(x,\cdot)$ is the $t$-step transition probability density for an initial state $x$. To analyze the inference time complexity, we now compute the number of MCMC iterations required until the distance is small enough. For a given fixed small $\epsilon$, the number of steps required for the distance to fall within $\epsilon$ is $t=\frac{\ln \epsilon}{\ln (1-\alpha)} = O(\frac{1}{\alpha})$.
Intuitively, the sampling algorithm without latent adaptation (Algorithm~\ref{alg:S2M_sampling}) takes time inversely proportional to the relative ratio of the target class samples out of the entire dataset used to train the generator.

Given a fixed number of data, as the number of attributes increases, $\alpha$ decreases, and the algorithm converges slowly. 
To alleviate such inefficiency, we additionally proposed latent adaptation which collects a certain amount of target class samples using Algorithm~\ref{alg:S2M_sampling} and then uses it fit to a new generator distribution close the target distribution. 
If the newly obtained generator distribution is sufficiently closed to the target distribution, the sampling algorithm using the new generator will take a constant time to produce the target class samples. The remaining question is how long it take to apply latent adaptation.

\input{Figures/Figure_Iterative_LA}

To discuss about the time complexity of latent adaptation scale with the number of attributes, we let $q_i$ be the probability of a generated sample belonging to the $i$-th class and assume conditional independence among the classes, \ie, $p(y_i| x, y_1,...,y_{i-1},y_{i+1},...,y_{n})=p(y_i| x)$. Let $m$ be the number of latent samples used to fit the new proposal distribution and $c$ be the overhead introduced by fitting the latent distribution. We consider two scenarios of applying latent adaptation and analyze the time complexity taken to completely fit the latent space in each scenario: (i) Applying latent adaptation by searching the latent space of target joint class samples at once. (ii) Applying latent adaptation repeatedly by increasing the number of specified attributes one by one as shown in Figure~\ref{fig:iterative_LA} (See Algorithm~\ref{alg:latent_adaptation_sharing_repeat} for details). For the case (i), the algorithm runs the MH algorithm once to draw the target class samples and the probability of each generated sample belonging to that class is $\prod_{i\in I} q_i \prod_{j\in J} (1-q_j)$. Hence, it takes $O(m(\prod_{i\in I} q_i \prod_{j\in J} (1-q_j))^{-1}+c)$. For the case (ii), the algorithm runs the MH algorithm once for each class of $I$ and each class of $J$, so it takes $O(m(\sum_{i\in I} q_i^{-1} + \sum_{j\in J} (1-q_j)^{-1})+c(\left|I\right|+\left|J\right|))$. That is, Algorithm~\ref{alg:latent_adaptation_sharing_repeat} can run efficiently as it takes time linearly proportional to the number of attributes. 
When the number of attributes is small, (i) is also a good way to use latent adaptation, since it has a small overhead of fitting the latent distribution. 

To validate the empirical effectiveness of applications of latent adaptation, we select three joint classes of CelebA-ABMNSW, whose ratio of training samples is lower than $3\%$. For the method (i), sampling for obtaining latent samples is performed by $90$ MCMC iterations. For the method (ii), the latent sampling is performed by $15$ MCMC iterations for each attribute. For both cases, we collect $10K$ latent samples to perform latent adaptation. After fitting a new latent distribution suit for a target joint class, we again perform a sampling algorithm to draw target joint class samples. We measure the number of MCMC iterations required until accuracy converges. As shown in Table~\ref{table:la_comparision}, latent adaptation drastically shortens the MCMC chain for all cases, and the method (ii) requires fewer MCMC iterations to draw the target joint class samples than the method (i).

\begin{table}[h!]
\begin{center}
\caption{Comparison of applications of latent adaptation on CelebA-ABMNSW. For each joint class, we denote the ratio of training samples of the class in the parentheses. "-" in LA type indicates that latent adaptation is not applied. \# of searching iterations refers to the total number of MCMC steps performed to fit the latent distribution. \# of sampling iterations refers to the number of MCMC steps required until accuracy converges. Accuracy and FID are measured at this step.}
\label{table:la_comparision}
\footnotesize
\begin{tabular}[\linewidth]{c | c | c c c c}
\toprule
Class (Ratio) & LA type & \makecell{\# of searching\\iterations} & \makecell{\# of sampling\\iterations} & Accuracy & FID \\

\midrule
\multirow{3}{*}{B+M+S-A-N-W (1.95\%)}& -
& 0
& 320
& 53.51\% 
& 44.72
 \\
&(i)
& 90
& 20
& 57.87\%
& 45.37
\\
&(ii)
& 90
& 15
& 59.08\%
& 46.41
\\

\midrule
\multirow{3}{*}{B+M+N+S-A-W (2.68\%)}& -
& 0
& 165
& 43.78\% 
& 38.64
 \\
&(i)
& 90
& 35
& 42.70\%
& 38.34
\\
&(ii)
& 90
& 5
& 44.14\%
& 37.15
\\

\midrule
\multirow{3}{*}{A+N+W-B-M-S (2.89\%)}& -
& 0
& 335
& 84.72\% 
& 32.03
 \\
&(i)
& 90
& 40
& 86.20\%
& 32.22
\\
&(ii)
& 90
& 5
& 86.94\% 
& 33.89
\\

\bottomrule
\end{tabular}
\end{center}
\end{table}

%% file: Tables/Table_running_time.tex
\begin{table}[!ht]
\begin{center}
\caption{Inference time (ms) of several architectures on CelebA-BMS. PD denote the projection discriminator inference, CLS denote the classifier inference, and GMM denote the gaussian mixture sampler inference. We use the batch size of 64.}
\label{table:running_time}
\footnotesize
\begin{tabular}[\linewidth]{l | c c c c c}
\toprule
Method &
CP-GAN & BigGAN & BigGAN + PD & BigGAN + CLS & BigGAN + CLS + GMM \\

\midrule

Time (ms) &
31.61 &
29.56 &
31.66 &
35.86 &
36.18
\\

\bottomrule
\end{tabular}
\end{center}
\end{table}

%% file: Figures/Figure_Iterative_LA.tex
\begin{figure}[t!]
\begin{center}
    \includegraphics[width=1.0\linewidth]{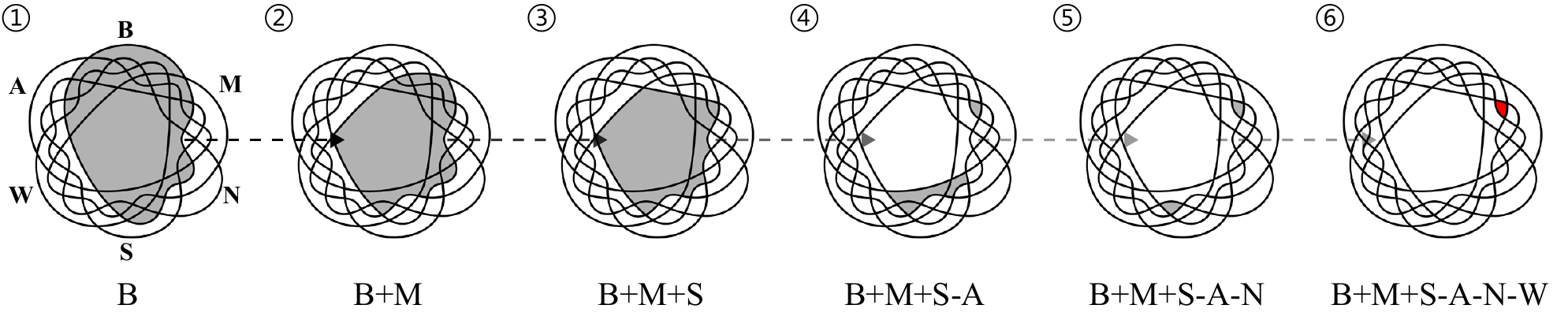}
\end{center}
    \caption{Our latent adaptation can be repeatedly applied for the efficiency of S2M sampling. For instance, to generate images of \emph{B+M+S-A-N-W} joint class (6) with S2M sampling, we can first search the space of \emph{Black-hair} attribute (1) and then gradually increase the number of attributes.}
\label{fig:iterative_LA}
\end{figure}

%% file: Appendix/D_Experiments_Details.tex
\section{Experiments Details}
\label{appendix:exp_details}

\subsection{MNIST and FMNIST}

Each of MNIST and FMNIST dataset consists of a training set of $60k$ images and a test set of $10k$ images. We use 10\% of the training set as the validation set. To make the training dataset annotated by single positive labels, we distribute the images belong to the joint classes equally to each corresponding classes of the single positive labels. We evaluate accuracy on MNIST and FMNIST dataset using LeNet5~\citep{726791} trained with fully annotated dataset.

The generative models for MNIST and FMNIST are discussed in Section~\ref{sec:exp_mnist}. As similar to the original setting of GenPU, the generator consists of ReLU activations and fully connected layers of input size: 100-256-256-784. The discriminator consists of ReLU activations and fully connected layers of input size: 784-256-256. 
As for the GAN objective, we follow the settings introduced by the authors for baselines, and use WGAN-GP~\citep{WeiGL0W18} for our model.
We train all generative models using Adam optimizer~\citep{KingmaB14} with a learning rate of $0.0001$, $\beta_1=0.5, \beta_2=0.999$, and a batch size of $64$. 
The generator is trained for $200k$ iterations, and two updates of the discriminator are performed for every update of the generator.

Classification networks used for S2M sampling are obtained from multiple branches of LeNet5 architecture. We train the classifier using Adam optimizer. For MNIST 3/5 dataset, the classifier is trained for 10 epochs with a learning rate of $0.001$, and the temperature of $D_r$ is set to 2. For MNIST and FMNIST Even dataset, the classifier is trained for 50 epochs with a learning rate of of $0.0001$, and the temperature of $D_v$ is set to 4. Each $\gamma_k$ corresponding to the intersection set is set to 0.1 for both MNIST and FMNIST.

\subsection{CIFAR-10 and CelebA}

For CIFAR-10 dataset, we use 10\% of the training set as the validation set. We follow the original partition description and resize images to $64 \times 64$ for training efficiency on CelebA dataset. To make the training datasets annotated by single positive labels, we distribute the images belong to the joint classes equally to each corresponding classes of the single positive labels. 
We evaluate accuracy on CIFAR-10 and CelebA dataset using MobileNet V2~\citep{SandlerHZZC18} trained with fully annotated dataset.

The generative models for CIFAR-7to3 and CelebA datasets are discussed in Section~\ref{sec:exp_real_data}. We use BigGAN~\citep{BrockDS19} architecture for all generative models and follow the PyTorch implementation\footnote{\url{https://github.com/POSTECH-CVLab/PyTorch-StudioGAN}}. 
We use hinge loss~\citep{LimY17} as the GAN objective and apply spectral normalization~\citep{MiyatoKKY18}. 
We train all models using Adam optimizer~\citep{KingmaB14} with a learning rate of $0.0002$, $\beta_1=0.5, \beta_2=0.999$, and a batch size of of $64$. 
The generator is trained for $100k$ iterations, and five updates of the discriminator are performed for every update of the generator. We select the model achieving best FID on the validation dataset.

Classification networks used for S2M sampling are obtained from multiple branches of MobileNet V2 architecture. We first train the classifier with only $\mathcal{L}_r$ during 200 epochs for CIFAR-7to3 and 30 epochs for CelebA-BMS, CelebA-ABMNSW dataset. We use SGD optimizer with a learning rate of $0.1$  and cosine annealing for this training. Then, the classifier is trained with the sum of all classification losses for $30k$ iterations. For CIFAR-7to3 and CelebA-BMS dataset, we set the temperature of $D_r$ and $D_f$ as 0.2, 1.0, and 1.2 when the size of difference index set is 0, 1, and 2, respectively. For CelebA-ABMNSW dataset, we set the temperature of $D_r$ and $D_f$ as 0.1, 0.1, 1.0, 1.0, 1.6 and 1.6 when the size of difference index set is 0, 1, 2, 3, 4, 5 and 6, respectively. Each $\gamma_k$ corresponding to the intersection set is set to 0.1 for CelebA-BMS and 0.5 for CelebA-ABMNSW. On CIFAR-7to3 dataset, we set 0.5 and 0.8 to this parameter for unconditional GAN and cGAN-PD, respectively.

\subsection{CelebA-HQ}

For CelebA-HQ dataset, we follow the original partition description and use the images with the resolution of $256\times 256$. To make the training dataset annotated by single positive labels (\ie, Black hair, Man and Smiling), we distribute the images belong to the joint classes equally to each corresponding classes of the single positive labels.

We use the pretrained StyleGAN V2~\citep{KarrasLAHLA20}. 
Classification networks used for S2M sampling are obtained from multiple branches of MobileNet V2~\citep{SandlerHZZC18} architecture. 
We first train the classifier with only $\mathcal{L}_r$ during 30 epochs. We use SGD optimizer with a learning rate of $0.1$  and cosine annealing for this training. Then, the classifier is trained with the sum of all classification losses for $1k$ iterations. We set the temperature of $D_r$ as 0.2, 1.0, and 1.2 when the size of difference index set is 0, 1, and 2, respectively. Each $\gamma_k$ corresponding to the intersection set is set to 0.5.

%% file: Appendix/E_Experimental_Results.tex
\section{Additional Experimental Results}
\label{appendix:exp_results}

\subsection{${2 \times 16}$ Gaussians Example}
\label{appendix:exp_gaussians}

\input{Figures/Figure_23Gaussian}

To provide an illustrative example, we modify \emph{25 Gaussians}~\citep{AzadiODGO19, TurnerHFSY19, CheZSLPCB20} to have two $4\times4$ grids of two-dimensional Gaussians of two classes $A$ and $B$ (See Figure~\ref{fig:exp_23_gaussian}). The 23 modes in ${2 \times 16}$ Gaussians are horizontally and vertically spaced by 1.0 and have a standard deviation of 0.05. 
We first train a GAN to randomly draw points within two grids using WGAN-GP~\citep{WeiGL0W18} as the GAN objective. To apply S2M sampling, we train the classifier to classify each point into two classes $A$ and $B$. The generator, discriminator, and classification networks consist of ReLU activations and fully connected layers of input size: 2-512-512-512. 

\input{Tables/Table_23Gaussian}

In this setting, S2M sampling attempts to draw points of given classes (A, B), the overlapping class ($A \cap B$), and the non-overlapping classes ($A \setminus B$, $B \setminus A$). We obtain samples at 400 MCMC iterations. 
As shown in Figure~\ref{fig:exp_23_gaussian}, S2M sampling correctly draw points of the target classes. On the other hand, GAN tends to generate spurious lines between the points. 
For the quantitative analysis, we report the accuracy, high-quality ratio, and mode standard deviation. We generate $10k$ samples and assign each point to the mode with the closest $L_2$ distance for measuring the accuracy. Following Turner \etal~\cite{TurnerHFSY19}, samples whose $L_2$ distances are within four standard deviations are considered as high-quality samples.
As shown in Table~\ref{table:exp_23_gaussians}, S2M sampling accurately draw samples for all conditions, and the ratio of high-quality samples is improved by $14.36\%$ on average.

\subsection{CelebA-HQ $256\times256$}

To validate whether S2M sampling can be adopted to state-of-the-art architecture on high resolution ($256\times256$) image dataset, we evaluate the performance of S2M sampling with the pretrained StyleGANv2~\cite{KarrasLAHLA20} on CelebA-HQ dataset.
The quantitative results for StyleGANv2 with and without S2M sampling are shown in Table~\ref{table:stylegan}, and the qualitative results are depicted in Section~\ref{appendix:more_results}.

\input{Tables/Table_StyleGAN}

\subsection{Data Embedding Visualization.}

To visually probe the effect of S2M sampling, 
we perform t-SNE~\citep{vanDerMaaten2008, PolicarSZ19} on generated samples. 
We embed Inception-V3 activations of samples generated by various models on CIFAR-7to3. As shown in Figure~\ref{fig:exp_tsne}, S2M sampling accurately draw samples for all classes compared to other generative models.

\input{Figures/Figure_TSNE}

\subsection{Ablation Study}
\label{appendix:ablation_study}

Since classifiers used in S2M sampling are not optimal in practice, we correct the sampling algorithm by scaling the temperatures of the classifiers and adjusting $\gamma_k$. Temperature scaling~\citep{GuoPSW17} is useful technique to control the confidence of classifier. If the temperature of $D_f$ and $D_r$ is small, samples that are likely to belong to overlapping classes tend to be drawn mostly in S2M sampling. Another approach is to adjust $\gamma_k$. We can obtain samples clearly distinct from classes of a difference set by decreasing $\gamma_i$ for all $i\in I$.

To validate the effects of the temperature scaling and $\gamma$ adjustment in S2M sampling, we perform the ablation studies on CIFAR-7to3 and CelebA-BMS dataset. The hyperparameter values used are explained in Appendix~\ref{appendix:exp_details}. In Table~\ref{table:ablation_study}, we report the average accuracy and FID when S2M sampling is applied to unconditional GAN and cGAN-PD. With the proper adjustment of the hyperparameters, accuracy can be highly improved while FID is maintained or slightly degraded.

\input{Tables/Table_Ablation_Study}

\subsection{Qualitative Results}
\label{appendix:more_results}

In this section, we provide the qualitative results for the experiments on CIFAR-7to3, CelebA-BMS, and CelebA-HQ. The qualitative results for CIFAR-7to3 and CelebA-BMS are shown in Figure~\ref{fig:exp_cifar10_baselines} and Figure~\ref{fig:exp_celeba_baselines}, respectively. 
One of advantages of S2M sampling is that it can be readily built upon existing state-of-the-art GANs. Figure~\ref{fig:exp_celebahq_qual} represents the samples with the resolution of $256\times 256$ when we adopted S2M sampling to pretrained StyleGAN V2~\citep{KarrasLAHLA20} on CelebA-HQ dataset.

\input{Figures/Figure_CIFAR-10_Baselines}
\input{Figures/Figure_CelebA_Baselines}
\input{Figures/Figure_CelebA_Sampling_256}

%% file: Figures/Figure_23Gaussian.tex
\begin{figure}[h!]
\begin{center}
    \includegraphics[width=1.0\linewidth]{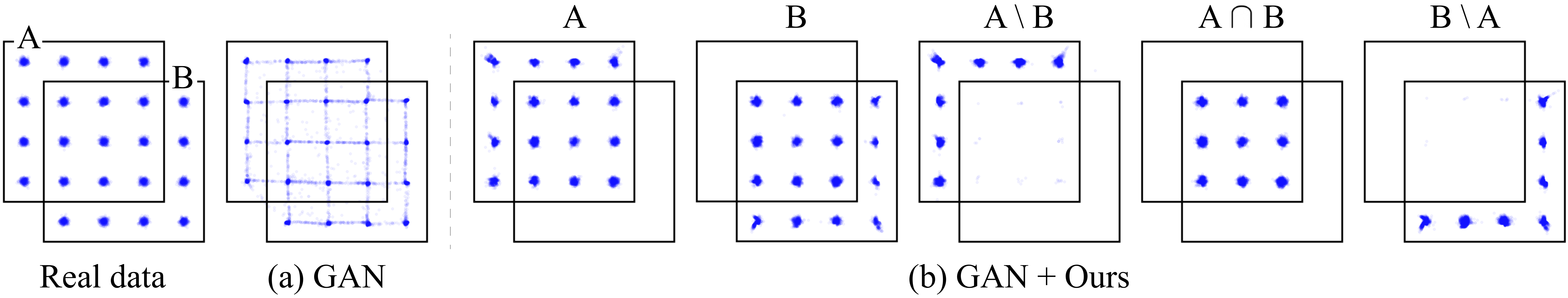}
\end{center}
\vspace{-0.1in}
    \caption{Example of $2 \times 16$ Gaussians. Using the outputs of original GAN, S2M Sampling samples high-quality points within various conditions ($\mathrm{A, B, A \setminus B, B \setminus A, A \cap B}$).}
\label{fig:exp_23_gaussian}
\end{figure}

%% file: Tables/Table_23Gaussian.tex
\begin{table}[!ht]
\caption{Accuracy (\%), high-quality ratio (\%), and mode standard deviation on $2 \times 16$ Gaussians.}
\label{table:exp_23_gaussians}
\centering
\footnotesize
\setlength\tabcolsep{5pt}

\begin{tabular}[\linewidth]{l c c c | c c c}
\toprule
        Condition & \multicolumn{3}{c}{GAN} & \multicolumn{3}{c}{\textbf{GAN + Ours}} \\
\midrule
                & Accuracy  & \makecell{High Quality} & \makecell{Std. Dev.} 
                & Accuracy  & \makecell{High Quality} & \makecell{Std. Dev.} \\
\midrule
$\mathrm{A, B}$    & \cell[69.83]{0.35}    & \cell[84.39]{0.60}    & \cell[0.106]{0.002}    & \cell[100.00]{0.00}   & \cell[98.94]{0.40}    & \cell[0.052]{0.002}     \\
$\mathrm{A \setminus B, B \setminus A}$        & \cell[30.17]{0.35}    & \cell[88.87]{0.50}    & \cell[0.090]{0.002}     & \cell[99.52]{0.36}    & \cell[98.67]{0.51}    & \cell[0.051]{0.002}     \\
$\mathrm{A \cap B}$     & \cell[39.66]{0.46}    & \cell[80.98]{0.82}    & \cell[0.118]{0.003}    & \cell[100.00]{0.00}   & \cell[99.73]{0.14}    & \cell[0.050]{0.001}     \\
\bottomrule
\end{tabular}
\end{table}

%% file: Tables/Table_StyleGAN.tex
\begin{table}[h!]
\begin{center}
\caption{Quantitative results of StyleGANv2 with S2M sampling.}
\label{table:stylegan}
\footnotesize
\begin{tabular}[\linewidth]{l | cc}
\toprule

Method & Accuracy ($\uparrow$) & FID ($\downarrow$) \\

\midrule

StyleGANv2 &
15.44\% &
17.08 \\

StyleGANv2 + \textbf{Ours} &
77.18\% &
14.64 \\
\bottomrule
\end{tabular}
\end{center}
\end{table}

%% file: Figures/Figure_TSNE.tex
\begin{figure}[h!]
\begin{center}
    \includegraphics[width=1.0\linewidth]{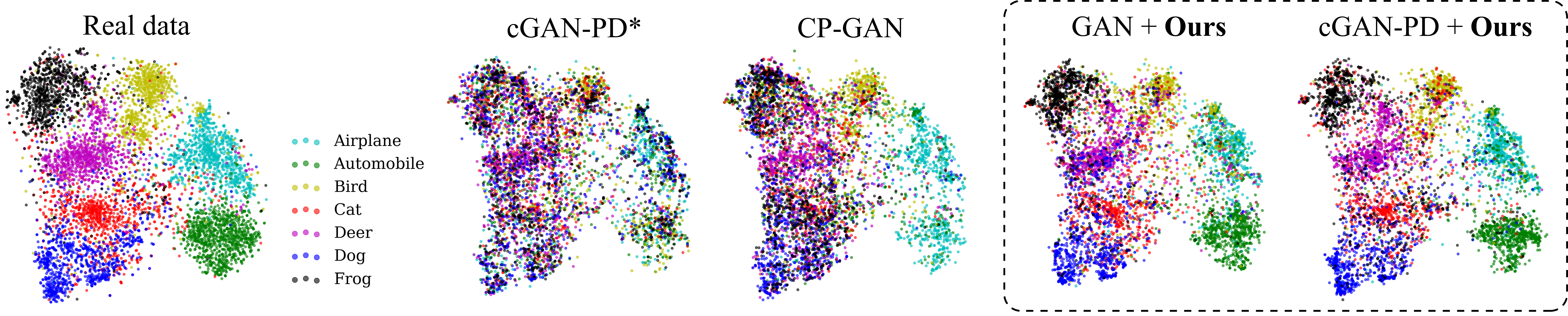}
\end{center}
\vspace{-0.1in}
\caption{t-SNE visualization results for generated samples on CIFAR-7to3 dataset. Samples drawn from S2M sampling are embedded similarly to the real data.
}
\vspace{-0.1in}
\label{fig:exp_tsne}
\end{figure}

%% file: Tables/Table_Ablation_Study.tex
\begin{table}[h!]
\caption{Ablation study for the hyperparameters of S2M sampling.}
\label{table:ablation_study}

\centering
\footnotesize

\setlength\tabcolsep{5pt}

\begin{tabular}[0.9\textwidth]{l c c c | c c}
\toprule
        \multirow{2}{*}{Method} & \multirow{2}{*}{Metric} & \multicolumn{2}{c}{CIFAR-7to3} & \multicolumn{2}{c}{CelebA-BMS} \\
        \cmidrule(lr){3-4} \cmidrule(lr){5-6}
        &   &   GAN & cGAN-PD
        &   GAN & cGAN-PD \\
\midrule
\multirow{2}{*}{\makecell[l]{Sampling w/\\actual logits}} & Acc. ($\uparrow$)
& \cell[62.73]{1.63} & \cell[72.33]{0.52} & \cell[71.14]{1.46} & \cell[73.32]{4.62} \\
& FID ($\downarrow$)
& \cell[14.99]{0.17} & \cell[14.11]{0.23} & \cell[9.86]{1.41} & \cell[10.29]{0.83} \\
\midrule

\multirow{2}{*}{\makecell[l]{+ temperature scaling}} & Acc. ($\uparrow$) 
& \cell[68.40]{0.82} & \cell[76.32]{1.44} & \cell[74.99]{1.95} & \cell[78.50]{2.63} \\
& FID ($\downarrow$)
& \cell[14.81]{0.23} & \cell[14.12]{0.31} & \cell[9.81]{1.21} & \cell[10.21]{0.41} \\
\midrule

\multirow{2}{*}{\makecell[l]{+ $\gamma$ adjustment}} & Acc. ($\uparrow$)
& \cell[77.65]{1.22} & \cell[80.62]{2.08} & \cell[85.22]{4.27} & \cell[90.44]{1.05} \\
& FID ($\downarrow$)
& \cell[14.42]{0.55} & \cell[14.14]{0.34} & \cell[10.50]{0.97} & \cell[10.63]{0.29} \\
\bottomrule
\end{tabular}
\end{table}

%% file: Figures/Figure_CIFAR-10_Baselines.tex
\begin{figure}[!ht]
\begin{center}
    \includegraphics[width=1.0\linewidth]{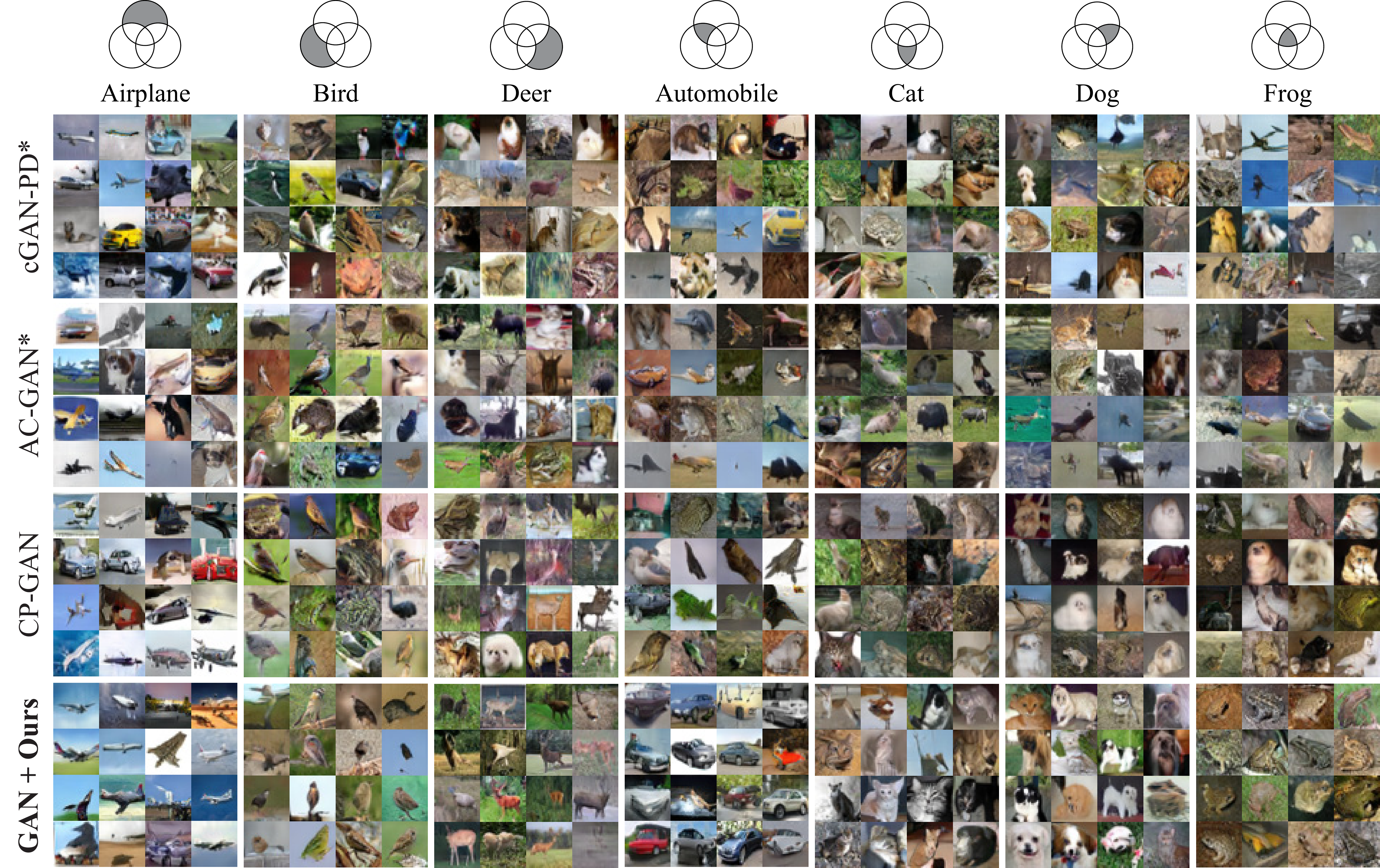}
\end{center}
    \caption{Qualitative results for $\textrm{cGAN-PD}^\ast$, $\textrm{AC-GAN}^\ast$, CP-GAN, and S2M sampling with GAN on CIFAR-7to3. For $\textrm{cGAN-PD}^\ast$, $\textrm{AC-GAN}^\ast$ and CP-GAN, we provide $1/n$ as the labels for each class to generate images belonging to $n$ classes.
    }
\label{fig:exp_cifar10_baselines}
\end{figure}

%% file: Figures/Figure_CelebA_Baselines.tex
\begin{figure}[!ht]
\begin{center}
    \includegraphics[width=1.0\linewidth]{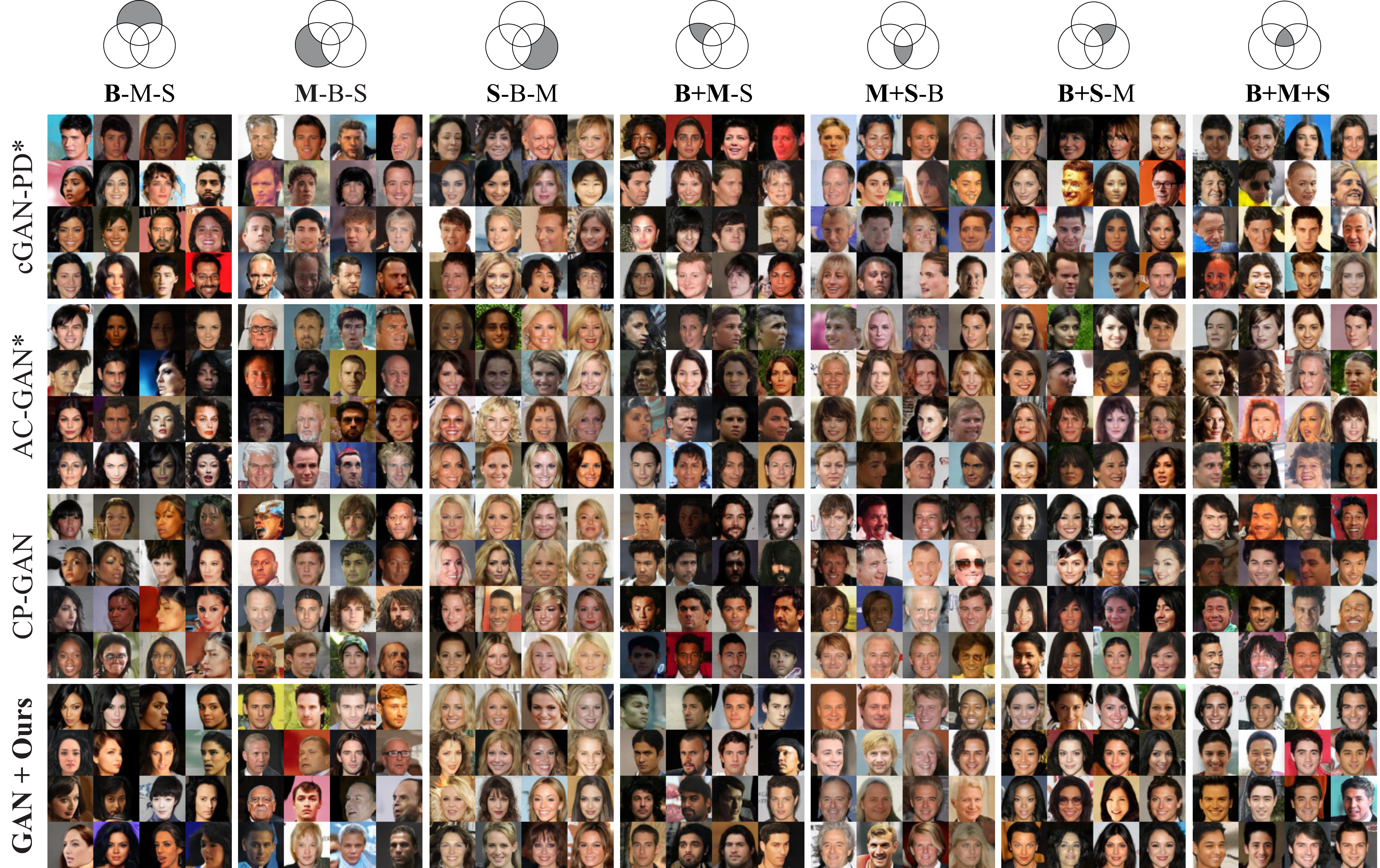}
\end{center}
    \caption{Qualitative results for $\textrm{cGAN-PD}^\ast$, $\textrm{AC-GAN}^\ast$, CP-GAN, and S2M sampling with GAN on CelebA-BMS. For $\textrm{cGAN-PD}^\ast$, $\textrm{AC-GAN}^\ast$ and CP-GAN, we provide $1/n$ as the labels for each class to generate images belonging to $n$ classes.
    Intersections and differences are denoted by plus signs and minus signs, respectively.}
\label{fig:exp_celeba_baselines}
\end{figure}

%% file: Figures/Figure_CelebA_Sampling_256.tex
\begin{figure}[!ht]
\begin{center}
    \includegraphics[width=1.0\linewidth]{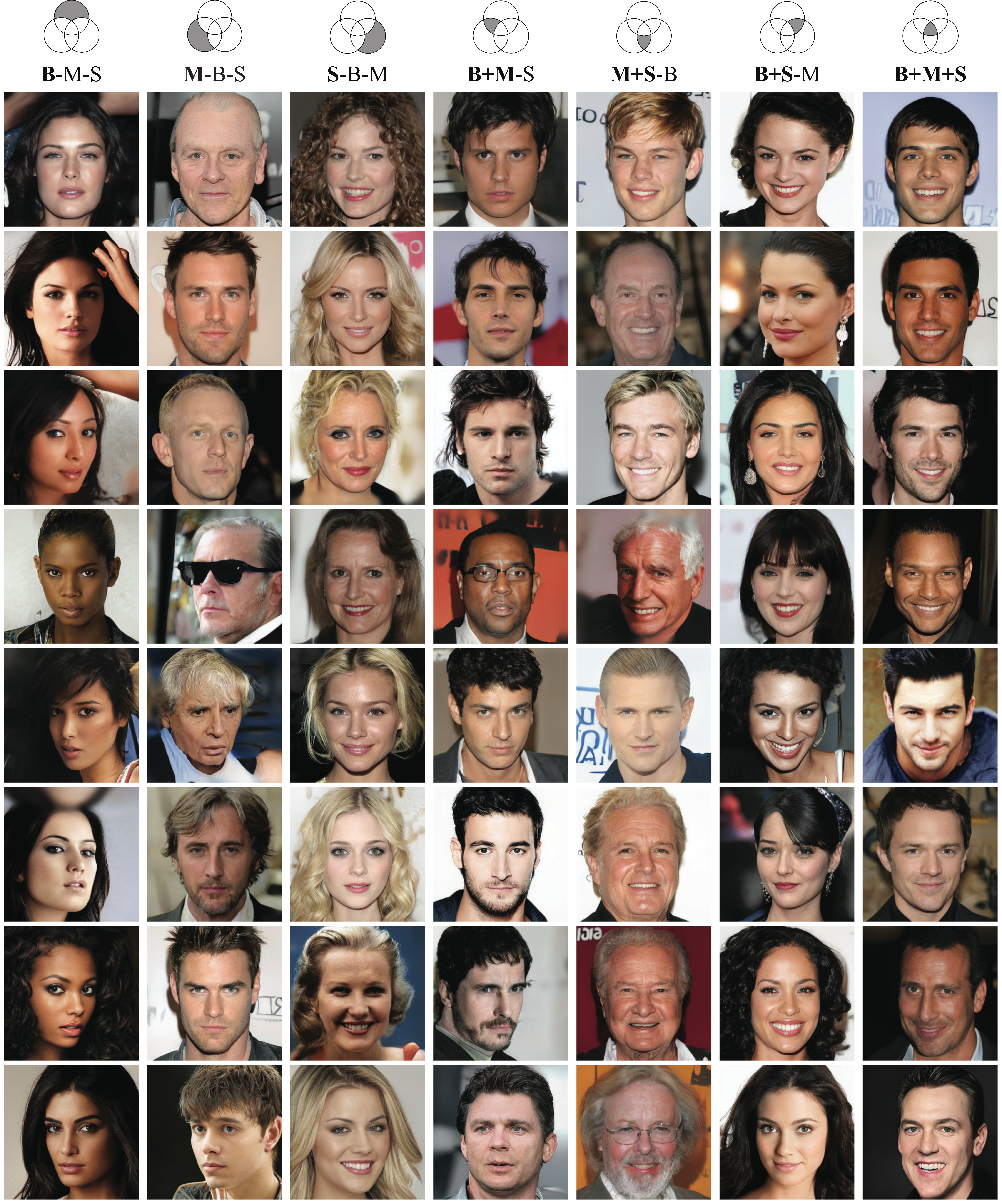}
\end{center}
\caption{Qualitative results of applying S2M sampling to StyleGANv2 on CelebA-HQ. Intersections and differences are denoted by plus signs and minus signs, respectively.}
\label{fig:exp_celebahq_qual}
\end{figure}

%% file: Appendix/F_Varying_Labels.tex
\section{S2M Sampling with Different Attributes}
\label{appendix:diff_attributes}

\input{Tables/Table_Other_Labels}
\input{Figures/Figure_CelebA_Other_Labels}

In this section, we examine S2M sampling with various attributes appearing in CelebA dataset. Concretely, we conduct two additional datasets: (i) CelebA-BBM consisting of classes of \emph{Brown-hair}, \emph{Bushy-eyebrows}, and \emph{Mouth-slightly-opens} attributes, (ii) CelebA-HBW consisting of classes of \emph{High-cheekbones}, \emph{Bags-under-eye}, and \emph{Wavy-hair} attributes.
Those attributes are chosen due to their strong visual impact. For both datasets, we adopt S2M sampling to pretrained StyleGANv2.

\textbf{Quantitative Results.} Table~\ref{table:other_labels} shows the quantitative results of S2M sampling. As shown in Table~\ref{table:other_labels}, we can observe that S2M sampling correctly draws samples corresponding to each joint class for both datasets, which indicates that S2M sampling can be efficiently performed with various types of attributes.

\textbf{Qualitative Results.} For the qualitative results, we depict samples drawn by S2M sampling for three joint classes of CelebA-BBM and CelebA-HBW in Figure~\ref{fig:other_labels}. For all joint classes, the samples in Figure~\ref{fig:other_labels} are randomly selected from the outputs of S2M sampling.

%% file: Tables/Table_Other_Labels.tex
\begin{table}[h!]
\begin{center}
\caption{Quantitative results of StyleGANv2 with S2M sampling on two variants of CelebA-BMS: CelebA-BBM and CelebA-HBW.}
\label{table:other_labels}
\vspace{0.1in}
\footnotesize
\begin{tabular}[\linewidth]{l | l cc}
\toprule

Dataset & Method & Accuracy ($\uparrow$) & FID ($\downarrow$) \\

\midrule\midrule

\multirow{2}{*}{CelebA-BBM} &
StyleGANv2 &
18.41\% &
15.57 \\

&
StyleGANv2 + \textbf{Ours} &
\textbf{74.97\%} &
\textbf{14.41} \\

\midrule

\multirow{2}{*}{CelebA-HBW} &
StyleGANv2 &
12.68\% &
15.80 \\

&
StyleGANv2 + \textbf{Ours} &
\textbf{70.94\%} &
\textbf{14.54} \\

\bottomrule
\end{tabular}
\end{center}
\vspace{-0.1in}
\end{table}

%% file: Figures/Figure_CelebA_Other_Labels.tex
\begin{figure}[!ht]
\begin{center}
    \includegraphics[width=0.85\linewidth]{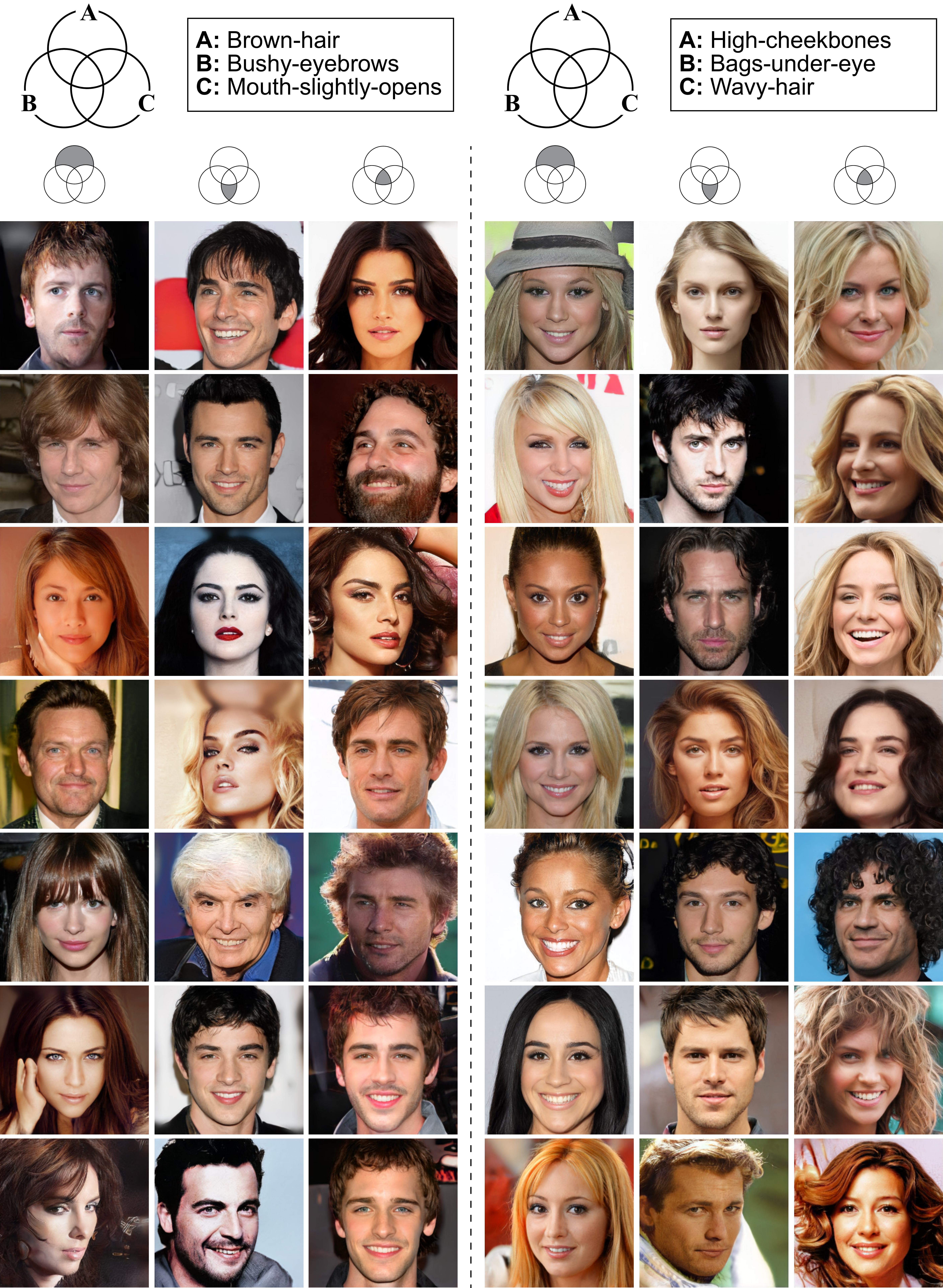}
\end{center}
\vspace{-0.1in}
\caption{Qualitative results of applying S2M sampling to StyleGANv2 on two varaints of CelebA-BMS: CelebA-BBM (left) and CelebA-HBW (right).
}
\label{fig:other_labels}
\end{figure}

%% file: Appendix/G_Consideration.tex
\section{Consideration of other sampling approaches for applying S2M Sampling}

In this section, we provide insight to apply S2M sampling with other sampling methods. We mainly considered applying three sampling algorithms which can be found in previous GAN sampling studies~\cite{AzadiODGO19, TurnerHFSY19, CheZSLPCB20, WangWYL21}; Rejection sampling, Independent Metropolis-Hastings algorithm, and Langevin dynamics. Here, we briefly discuss the pros and cons of each sampling method we faced in our problem settings.

\textbf{Rejection Sampling} As discussed in DRS~\cite{AzadiODGO19}, the rejection sampling can be applied to sample from the target distribution $p_t(x)$ if we can compute the ratio between the target density $p_t(x)$ and the generator density $p_G(x)$, and the upper bound of the density ratio $p_t(x) / p_G(x)$. However, it is in general difficult and expensive to compute this upper bound in the high dimensional data space. We need several heuristics to mitigate the issues, e.g., shifting the logit score of the classifier used to compute the density ratio as introduced by DRS~\cite{AzadiODGO19}, which may introduce additional non-trivial efforts for hyperparameter searching.

\textbf{Independent Metropolis-Hastings algorithm} This algorithm can also be used to draw samples from the target distribution if we can compute the density ratio $p_t(x) / p_G(x)$. Unlike Rejection sampling, we do not need to compute the upper bound of this density ratio, but a sequence of samples forming a Markov chain is required. Our study mainly used this algorithm because it empirically performed well without complex heuristics and the sampling accuracy can be readily controlled at the cost of MCMC steps. To further mitigate the sample efficiency in our problem, we suggest the latent adaptation technique.

\textbf{Langevin dynamics} Langevin dynamics is a gradient-based MCMC approach which can also be used when we can compute the density ratio $p_t(x) / p_G(x)$. Several studies~\cite{CheZSLPCB20, WangWYL21} employ its Euler-Maruyama discretization to improve the quality of GAN samples. While this algorithm can efficiently push a chain of samples towards the target distribution, the step size of the algorithm is very sensitive to the sampling cost and quality. Especially, in our problem, we need to deal with the case that the sample falls within the space where the gradient is not well-defined, i.e. $\supp p_G \setminus \supp p_t$. We did not use the algorithm as we could not find an effective way to address these issues.

%% file: Appendix/Social_Impacts.tex
\section{Potential Societal Impacts}

This work demonstrates that it is possible to generate multi-label data from limited labels.
In addition, this work can be freely adopted to unconditional GANs trained with a large amount of unlabeled data.
Hence, our work can reduce the high annotation cost that research groups face in common.
Despite the fact that deep learning models tend to struggle from learning underrepresented data~\citep{abs-1908-09635}, 
properly calibrated sampling algorithm does not readily ignore rarely appearing data, meaning that it is unlikely to introduce bias into generative models.